\documentclass[11pt,a4paper]{article}
\usepackage[hyperref]{acl2017}
\usepackage{times}
\usepackage{latexsym}
\usepackage{amsmath}
\usepackage{breqn}
\usepackage{pgfplotstable}
\usepackage{algorithm2e}
\usepackage{hhline}
\usepackage{multirow}
\usepackage{multicol}
\usepackage[font=small]{caption}
\usepackage{subcaption}
\usepackage{color}
\usepackage{float}
\usepackage{lipsum,adjustbox}
\usepackage{tikz}
\usepackage{tikz-dependency}
\usepackage{enumitem}
\usepackage{xr}

\usetikzlibrary{shapes,fit,calc,er,positioning,intersections,decorations.shapes,mindmap,trees}
\tikzset{decorate sep/.style 2 args={decorate,decoration={shape backgrounds,shape=circle,
      shape size=#1,shape sep=#2}}}

\newcommand{\com}[1]{}
\newcommand{\parser}[1]{TUPA\textsubscript{#1}}
\newcommand{\secref}[1]{Section~\ref{#1}}
\newcommand{\figref}[1]{Figure~\ref{#1}}
\newcommand{\tabref}[1]{Table~\ref{#1}}

\SetKwRepeat{Do}{do}{while}

\def\xlinkspace#1 #2{%
 \ifx\relax#2%
 \xlinkdash#1-\relax
 \else
 \xlinkdash#1 -\relax
 \expandafter\xlinkspace\expandafter#2%
 \fi}

\def\xlinkdash#1-#2{%
 \ifx\relax#2%
 \tmp{#1}%
 \else
 \tmp{#1-}%
 \expandafter\xlinkdash\expandafter#2%
 \fi}

\aclfinalcopy 
\def\aclpaperid{193} 

\title{A Transition-Based Directed Acyclic Graph Parser for UCCA}

\author{Daniel Hershcovich$^{1,2}$ \\
  \\\And
  Omri Abend$^2$ \\
  $^1$The Edmond and Lily Safra Center for Brain Sciences \\
  $^2$School of Computer Science and Engineering \\
  Hebrew University of Jerusalem \\
  \texttt{\{danielh,oabend,arir\}@cs.huji.ac.il}
  \\\And
  Ari Rappoport$^2$
}

\date{}

\begin{document}
\maketitle

\begin{abstract}
  We present the first parser for UCCA, a
  cross-linguistically applicable framework for semantic
  representation, which builds on extensive
  typological work and supports rapid annotation.
  UCCA poses a challenge for existing parsing techniques,
  as it exhibits reentrancy (resulting in DAG structures),
  discontinuous structures and non-terminal nodes corresponding
  to complex semantic units. To our knowledge, the conjunction
  of these formal properties is not supported by any existing parser.
  Our transition-based parser, which uses a novel transition set
  and features based on bidirectional LSTMs,
  has value not just for UCCA parsing:
  its ability to handle more general graph structures can inform
  the development of parsers for other semantic DAG structures, 
  and in languages that frequently use discontinuous structures.
\end{abstract}

\section{Introduction}\label{sec:introduction}

Universal Conceptual Cognitive Annotation \cite[UCCA,][]{abend2013universal}
is a cross-linguistically applicable semantic representation scheme,
building on the established Basic Linguistic Theory typological framework
\cite{Dixon:10b,Dixon:10a,Dixon:12}, and Cognitive
Linguistics literature \cite{croft2004cognitive}.
It has demonstrated applicability to multiple languages, including
English, French, German and Czech,
support for rapid annotation by non-experts
(assisted by an accessible annotation interface \cite{abend2017uccaapp}),
and stability under translation \cite{sulem2015conceptual}.
It has also proven useful for machine translation evaluation \cite{birch2016hume}.
UCCA differs from syntactic
schemes in terms of content and formal structure.
It exhibits reentrancy, discontinuous nodes and non-terminals,
which no single existing parser supports.
Lacking a parser, UCCA's applicability has been so far limited,
a gap this work addresses.

We present the first UCCA parser, \parser{}
(Transition-based UCCA Parser),
building on recent advances in discontinuous constituency
and dependency graph parsing, and further introducing novel transitions and features for UCCA.
Transition-based techniques are a natural
starting point for UCCA parsing, given the conceptual similarity of
UCCA's distinctions, centered around predicate-argument structures, to distinctions expressed
by dependency schemes, and the achievements of transition-based methods in dependency parsing
\cite{dyer2015transition,andor2016globally,kiperwasser2016simple}.
We are further motivated by the strength of transition-based methods
in related tasks, including dependency graph parsing
\cite{sagae2008shift,ribeyre-villemontedelaclergerie-seddah:2014:SemEval,tokgoz2015transition},
constituency parsing \cite{sagae2005classifier,zhang2009transition,zhu2013fast,maier2015discontinuous,maier-lichte:2016:DiscoNLP},
AMR parsing \cite{wang-xue-pradhan:2015:ACL-IJCNLP,wang2015transition,wang-EtAl:2016:SemEval,dipendra2016neural,goodman2016noise,zhou2016amr,damonte-17}
and CCG parsing \cite{zhang2011shift,ambati2015incremental,ambati-deoskar-steedman:2016:N16-1}.

We evaluate \parser{} on the English UCCA corpora, including in-domain and out-of-domain settings.
To assess the ability of existing
parsers to tackle the task, we develop a conversion procedure
from UCCA to bilexical graphs and trees.
Results show superior performance for \parser{}, demonstrating the effectiveness of
the presented approach.\footnote{All parsing and conversion code, as well as trained parser models,
are available at \url{https://github.com/danielhers/tupa}.}

The rest of the paper is structured as follows:
\secref{sec:ucca} describes UCCA in more detail.
\secref{sec:parser} introduces \parser{}.
\secref{sec:exp_setup} discusses the data and experimental setup.
\secref{sec:results} presents the experimental results.
\secref{sec:related_work} summarizes related work, and
\secref{sec:conclusion} concludes the paper.

\section{The UCCA Scheme}\label{sec:ucca}

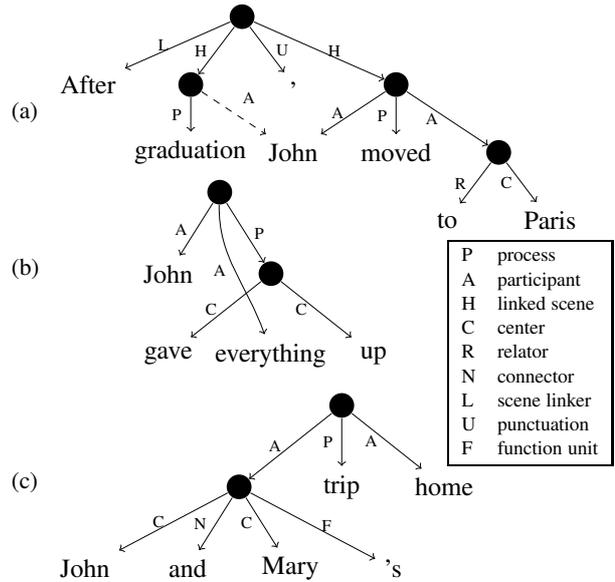
\begin{figure}[t]
  \begin{subfigure}{.9\columnwidth}
  \parbox{.05\columnwidth}{\caption{}\label{fig:graduation}}
  \parbox{.8\columnwidth}{
  \scalebox{.9}{
  \begin{tikzpicture}[level distance=10mm, ->,
      every circle node/.append style={fill=black}]
    \node (ROOT) [circle] {}
      child {node (After) {After} edge from parent node[left] {\scriptsize L}}
      child {node (graduation) [circle] {}
      {
        child {node {graduation} edge from parent node[left] {\scriptsize P}}
      } edge from parent node[left] {\scriptsize H} }
      child {node {,} edge from parent node[right] {\scriptsize U}}
      child {node (moved) [circle] {}
      {
        child {node (John) {John} edge from parent node[left] {\scriptsize A}}
        child {node {moved} edge from parent node[left] {\scriptsize P}}
        child {node [circle] {}
        {
          child {node {to} edge from parent node[left] {\scriptsize R}}
          child {node {Paris} edge from parent node[left] {\scriptsize C}}
        } edge from parent node[left] {\scriptsize A} }
      } edge from parent node[right] {\scriptsize H} }
      ;
    \draw[dashed,->] (graduation) to node [auto] {\scriptsize A} (John);
  \end{tikzpicture}
  }}
  \end{subfigure}
  \begin{subfigure}{.9\columnwidth}
  \vspace{-2cm}
  \parbox{.05\columnwidth}{\caption{}\label{fig:gave}}
  \parbox{.75\columnwidth}{\hspace{1cm}
  \scalebox{.9}{
  \begin{tikzpicture}[level distance=12mm, ->,
      every node/.append style={midway},
      every circle node/.append style={fill=black}]
    \node (ROOT) [circle] {}
      child {node {John} edge from parent node[left] {\scriptsize A}}
      child {node [circle] {}
      {
      	child {node {gave} edge from parent node[left] {\scriptsize C}}
      	child {node (everything) {everything} edge from parent[white]}
      	child {node {up} edge from parent node[left] {\scriptsize C}}
      } edge from parent node[right] {\scriptsize P} }
      ;
    \draw[bend right,->] (ROOT) to[out=-20, in=180] node [left] {\scriptsize A} (everything);
  \end{tikzpicture}
  }}
  \parbox{.1\columnwidth}{
  \vspace{2cm}
  \begin{adjustbox}{width=.3\columnwidth,margin=1pt,frame}
  \begin{tabular}{ll}
	  P & process \\
	  A & participant \\
	  H & linked scene \\
	  C & center \\
	  R & relator \\
	  N & connector \\
	  L & scene linker \\
	  U & punctuation \\
	  F & function unit
  \end{tabular}
  \end{adjustbox}
  }
  \end{subfigure}
  \begin{subfigure}{.9\columnwidth}
  \vspace{-1cm}
  \parbox{.05\columnwidth}{\caption{}\label{fig:home}}
  \parbox{.65\columnwidth}{
  \scalebox{.9}{
  \begin{tikzpicture}[level distance=12mm, ->,
      every node/.append style={midway},
      every circle node/.append style={fill=black}]
    \node (ROOT) [circle] {}
      child {node [circle] {}
      {
        child {node {John} edge from parent node[left] {\scriptsize C}}
        child {node {and} edge from parent node[left] {\scriptsize N}}
        child {node {Mary} edge from parent node[left] {\scriptsize C}}
        child {node {'s} edge from parent node[right] {\scriptsize F}}
      } edge from parent node[left] {\scriptsize A} }
      child {node {trip} edge from parent node[left] {\scriptsize P}}
      child {node {home} edge from parent node[left] {\scriptsize A}}
      ;
  \end{tikzpicture}
  }}
  \end{subfigure}
  \caption{\label{fig:examples}
    UCCA structures demonstrating three structural properties exhibited by
    the scheme.
    (\subref{fig:graduation}) includes a remote edge (dashed),
    resulting in ``John'' having two parents.
    (\subref{fig:gave}) includes a discontinuous unit (``gave ... up'').
    (\subref{fig:home}) includes a coordination construction (``John and Mary'').
    Pre-terminal nodes are omitted for brevity.
    Right: legend of edge labels.
  }
\end{figure}

UCCA graphs are labeled, directed acyclic graphs (DAGs),
whose leaves correspond to the tokens of
the text. A node (or {\it unit}) corresponds to a terminal or
to several terminals (not necessarily contiguous) viewed as a
single entity according to semantic or cognitive considerations.
Edges bear a category, indicating the role of the sub-unit in the parent relation.
\figref{fig:examples} presents a few examples.

UCCA is a multi-layered representation, where each layer corresponds
to a ``module'' of semantic distinctions.
UCCA's \textit{foundational layer}, targeted in this paper, covers the predicate-argument
structure evoked by predicates of all grammatical categories
(verbal, nominal, adjectival and others), the inter-relations between them,
and other major linguistic phenomena such as coordination and multi-word expressions.
The layer's basic notion is the \textit{scene},
describing a state, action, movement or some other relation that evolves in time.
Each scene contains one main relation (marked as either a Process or a State),
as well as one or more Participants.
For example, the sentence ``After graduation, John moved to Paris'' (\figref{fig:graduation})
contains two scenes, whose main relations are ``graduation'' and ``moved''.
``John'' is a Participant in both scenes, while ``Paris'' only in the latter.
Further categories account for inter-scene relations and the internal structure of
complex arguments and relations (e.g. coordination, multi-word expressions and modification).

One incoming edge for each non-root node is marked as \textit{primary},
and the rest (mostly used for implicit relations and arguments) as \textit{remote} edges,
a distinction made by the annotator.
The primary edges thus form a tree structure, whereas the remote edges enable reentrancy,
forming a DAG.

While parsing technology in general, and transition-based parsing in particular,
is well-established for syntactic parsing, UCCA
has several distinct properties
that distinguish it from syntactic representations,
mostly UCCA's tendency to abstract away from syntactic detail that do not
affect argument structure.
For instance, consider the following examples where the concept of a scene
has a different rationale from the syntactic concept of a clause.
First, non-verbal predicates in UCCA are represented like verbal ones,
such as when they appear in copula clauses or noun phrases. Indeed,
in \figref{fig:graduation}, ``graduation'' and ``moved'' are considered separate events,
despite appearing in the same clause. 
Second, in the same example, ``John'' is marked as a (remote) Participant
in the graduation scene, despite not being overtly marked.
Third, consider the possessive construction in \figref{fig:home}.
While in UCCA ``trip'' evokes a scene in which ``John and Mary'' is
a Participant, a syntactic scheme would analyze this phrase similarly to ``John and Mary's shoes''.

These examples demonstrate that a UCCA parser, and more generally semantic parsers,
face an additional level of ambiguity compared to their syntactic counterparts
(e.g., ``after graduation'' is formally very similar to ``after 2pm'',
which does not evoke a scene).
\secref{sec:related_work} discusses UCCA in the context of other semantic schemes,
such as AMR \cite{banarescu2013abstract}.


Alongside recent progress in dependency parsing into projective trees,
there is increasing interest in parsing into 
representations with more general structural properties  (see \secref{sec:related_work}).
One such property is \textit{reentrancy},
namely the sharing of semantic units between predicates.
For instance, in \figref{fig:graduation},
``John'' is an argument of both ``graduation''
and ``moved'', yielding a DAG rather than a tree.
A second property is \textit{discontinuity},
as in \figref{fig:gave}, where ``gave up'' forms a discontinuous semantic unit.
Discontinuities are pervasive, e.g.,  with multi-word
expressions \cite{schneider2014discriminative}.
Finally, unlike most dependency schemes, UCCA uses \textit{non-terminal nodes}
to represent units comprising more than one word.
The use of non-terminal nodes is motivated by constructions with no clear head, including
coordination structures (e.g., ``John and Mary'' in \figref{fig:home}),
some multi-word expressions (e.g., ``The Haves and the \textit{Have Nots}''),
and prepositional phrases (either the preposition or the head noun can serve as the constituent's head).
To our knowledge, no existing parser supports all structural properties required for UCCA
parsing.

\section{Transition-based UCCA Parsing}\label{sec:parser}

We now turn to presenting \parser{}.
Building on previous work on parsing reentrancies, discontinuities and non-terminal nodes,
we define an extended set of transitions and features that supports the conjunction of
these properties.

Transition-based parsers \cite{Nivre03anefficient} scan the text from start to end,
and create the parse incrementally by applying a \textit{transition}
at each step to the parser's state,
defined using three data structures: a buffer $B$ of tokens and nodes to be processed,
a stack $S$ of nodes currently being processed,
and a graph $G=(V,E,\ell)$ of constructed nodes and edges,
where $V$ is the set of \emph{nodes}, $E$ is the set of \emph{edges},
and $\ell : E \to L$ is the \emph{label} function, $L$ being the set of possible labels.
Some states are marked as \textit{terminal}, meaning that $G$ is the final output.
A classifier is used at each step to select the next transition based on features
encoding the parser's current state.
During training, an oracle creates training instances for the classifier,
based on gold-standard annotations.

\begin{figure*}
	\begin{adjustbox}{width=\textwidth,margin=3pt,frame}
	\begin{tabular}{llll|l|llllc|c}
		\multicolumn{4}{c|}{\textbf{\small Before Transition}} & \textbf{\small Transition} & \multicolumn{5}{c|}{\textbf{\small After Transition}} & \textbf{\small Condition} \\
		\textbf{\footnotesize Stack} & \textbf{\footnotesize Buffer} & \textbf{\footnotesize Nodes} & \textbf{\footnotesize Edges} & & \textbf{\footnotesize Stack} & \textbf{\footnotesize Buffer} & \textbf{\footnotesize Nodes} & \textbf{\footnotesize Edges} & \textbf{\footnotesize Terminal?} & \\
		$S$ & $x \;|\; B$ & $V$ & $E$ & \textsc{Shift} & $S \;|\; x$ & $B$ & $V$ & $E$ & $-$ & \\
		$S \;|\; x$ & $B$ & $V$ & $E$ & \textsc{Reduce} & $S$ & $B$ & $V$ & $E$ & $-$ & \\
		$S \;|\; x$ & $B$ & $V$ & $E$ & \textsc{Node$_X$} & $S \;|\; x$ & $y \;|\; B$ & $V \cup \{ y \}$ & $E \cup \{ (y,x)_X \}$ & $-$ &
		$x \neq \mathrm{root}$ \\
		$S \;|\; y,x$ & $B$ & $V$ & $E$ & \textsc{Left-Edge$_X$} & $S \;|\; y,x$ & $B$ & $V$ & $E \cup \{ (x,y)_X \}$ & $-$ &
		\multirow{4}{50pt}{\vspace{-5mm}\[\left\{\begin{array}{l}
		x \not\in w_{1:n},\\
		y \neq \mathrm{root},\\
		y \not\leadsto_G x
		\end{array}\right.\]} \\
		$S \;|\; x,y$ & $B$ & $V$ & $E$ & \textsc{Right-Edge$_X$} & $S \;|\; x,y$ & $B$ & $V$ & $E \cup \{ (x,y)_X \}$ & $-$ & \\
		$S \;|\; y,x$ & $B$ & $V$ & $E$ & \textsc{Left-Remote$_X$} & $S \;|\; y,x$ & $B$ & $V$ & $E \cup \{ (x,y)_X^* \}$ & $-$ & \\
		$S \;|\; x,y$ & $B$ & $V$ & $E$ & \textsc{Right-Remote$_X$} & $S \;|\; x,y$ & $B$ & $V$ & $E \cup \{ (x,y)_X^* \}$ & $-$ & \\
		$S \;|\; x,y$ & $B$ & $V$ & $E$ & \textsc{Swap} & $S \;|\; y$ & $x \;|\; B$ & $V$ & $E$ & $-$ &
		$\mathrm{i}(x) < \mathrm{i}(y)$ \\
		$[\mathrm{root}]$ & $\emptyset$ & $V$ & $E$ & \textsc{Finish} & $\emptyset$ & $\emptyset$ & $V$ & $E$ & $+$ & \\
	\end{tabular}
	\end{adjustbox}
	\caption{\label{fig:transitions}
	  The transition set of \parser{}. 
	  We write the stack with its top to the right and the buffer with its head to the left.
	  $(\cdot,\cdot)_X$ denotes a primary $X$-labeled edge, and $(\cdot,\cdot)_X^*$ a remote $X$-labeled edge.
	  $\mathrm{i}(x)$ is a running index for the created nodes.
	  In addition to the specified conditions,
	  the prospective child in an \textsc{Edge} transition must not already have a primary parent.
	}
\end{figure*}

\paragraph{Transition Set.}
Given a sequence of tokens $w_1, \ldots, w_n$, we predict a UCCA graph $G$ over the sequence.
Parsing starts with a single node on the stack (an artificial root node), and the input tokens
in the buffer. \figref{fig:transitions} shows the transition set.

In addition to the standard \textsc{Shift} and \textsc{Reduce} operations, 
we follow previous work in transition-based constituency parsing \cite{sagae2005classifier},
adding the \textsc{Node} transition for creating new non-terminal nodes.
For every $X\in L$,
\textsc{Node$_X$} creates a new node on the buffer as a parent of the first element on the stack, with an $X$-labeled edge.
\textsc{Left-Edge$_X$} and \textsc{Right-Edge$_X$} create a new primary $X$-labeled edge between the first two elements on the stack, where the parent is the left or the right node, respectively.
As a UCCA node may only have one incoming primary edge,
\textsc{Edge} transitions are disallowed if the child node already
has an incoming primary edge.
\textsc{Left-Remote$_X$} and \textsc{Right-Remote$_X$} do not have this restriction,
and the created edge is additionally marked as \textit{remote}.
We distinguish between these two pairs of transitions to allow the parser to create remote edges
without the possibility of producing invalid graphs.
To support the prediction of multiple parents, node and edge transitions
leave the stack unchanged, as in other work on
transition-based dependency graph parsing
\cite{sagae2008shift,ribeyre-villemontedelaclergerie-seddah:2014:SemEval,tokgoz2015transition}.
\textsc{Reduce} pops the stack, to allow removing a node
once all its edges have been created.
To handle discontinuous nodes, \textsc{Swap} pops the second
node on the stack and adds it to the top of the buffer, as with the similarly
named transition in previous work \cite{nivre2009non,maier2015discontinuous}.
Finally, \textsc{Finish} pops the root node and marks the state as terminal.

\paragraph{Classifier.}
The choice of classifier and feature representation has been shown to play an important role in
transition-based parsing \cite{chen2014fast,andor2016globally,kiperwasser2016simple}.
To investigate the impact of the type of transition classifier in UCCA parsing,
we experiment with three different models.
\begin{enumerate}[leftmargin=*]
\item
Starting with a simple and common choice \cite[e.g.,][]{maier-lichte:2016:DiscoNLP},
\textbf{\parser{Sparse}} uses a linear classifier with sparse features, trained with
the averaged structured perceptron algorithm
\cite{Coll:04} and \textsc{MinUpdate} \cite{goldberg2011learning}:
each feature requires a minimum number of updates in training
to be included in the model.\footnote{We also experimented with a linear model using
dense embedding features, trained with the averaged structured perceptron algorithm.
It performed worse than the sparse perceptron model and was hence discarded.}
\item
Changing the model to a feedforward neural network with dense embedding features,
\textbf{\parser{MLP}}
(``multi-layer perceptron''), uses an architecture similar to that of \citet{chen2014fast},
but with two rectified linear layers instead of one layer with cube activation.
The embeddings and classifier are trained jointly.
\item
Finally, \textbf{\parser{BiLSTM}} uses a bidirectional LSTM for feature representation,
on top of the dense embedding features,
an architecture similar to \citet{kiperwasser2016simple}.
The BiLSTM runs on the input tokens in forward and backward directions,
yielding a vector representation that is then concatenated with dense features representing the
parser state (e.g., existing edge labels and previous parser actions; see below).
This representation is then fed into a feedforward network similar to \parser{MLP}.
The feedforward layers, BiLSTM and embeddings are all trained jointly.
\end{enumerate}

For all classifiers, inference is performed greedily, i.e., without beam search.
Hyperparameters are tuned on the development set (see \secref{sec:exp_setup}).

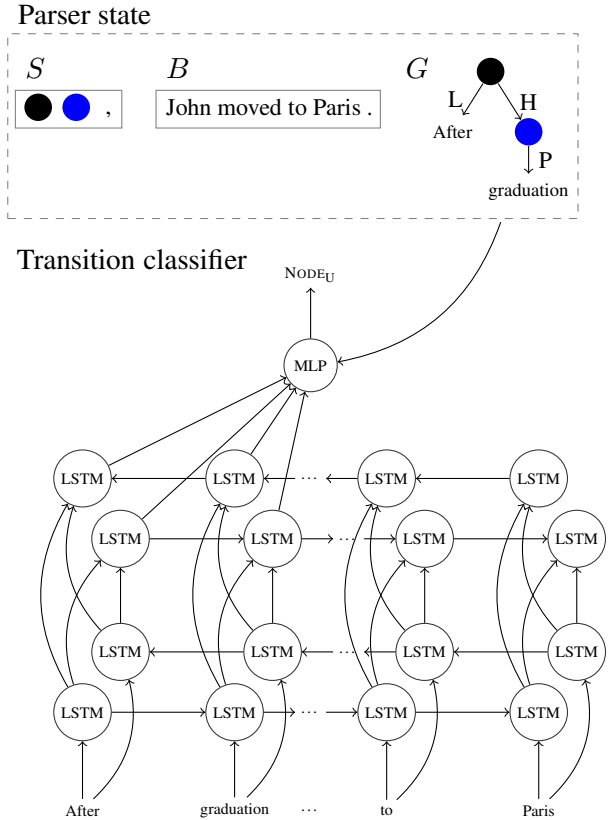
\begin{figure}[t]
	\begin{tikzpicture}[level distance=8mm, sibling distance=1cm]
	\node[anchor=west] at (0,1.5) {Parser state};
	\draw[color=gray,dashed] (0,-1.2) rectangle (7.5,1.25);
	\draw[color=gray] (.1,0) rectangle (1.5,.5);
	\node[anchor=west] at (.1,.8) {$S$};
	\node[fill=black, circle] at (.4,.275) {};
	\node[fill=blue, circle] at (.9,.275) {};
	\node[anchor=west] at (1.15,.175) {\small ,};
	\draw[color=gray] (1.95,0) rectangle (4.9,.5);
	\node[anchor=west] at (1.95,.8) {$B$};
	\node[anchor=west] at (1.95,.275) {\small John moved to Paris .};
	\node[anchor=west] at (5.1,.8) {$G$};
	\node[fill=black, circle] at (6.35,.75) {}
	  child {node  {\scriptsize After} edge from parent [->] node[left] {\small L}}
	  child {node [fill=blue, circle] {}
	  {
	    child {node {\scriptsize graduation} edge from parent [->] node[right] {\small P}}
	  } edge from parent [->] node[right] {\small H} };
	\end{tikzpicture}
	\begin{tikzpicture}[->]
	\node[anchor=west] at (0,6) {Transition classifier};
	\tiny
	\tikzstyle{main}=[circle, minimum size=7mm, draw=black!80, node distance=12mm]
	\foreach \i/\word in {1/{After},3/{graduation},5/{to},7/{Paris}} {
	    \node (x\i) at (\i,-1.3) {\word};
	    \node[main, fill=white!100] (h\i) at (\i,0) {LSTM};
        \path (x\i) edge (h\i);
	    \node[main, fill=white!100] (i\i) at (\i.5,.8) {LSTM};
        \path (x\i) edge [bend right] (i\i);
	    \node[main, fill=white!100] (l\i) at (\i.5,2.3) {LSTM};
        \path (h\i) edge [bend left] (l\i);
        \path (i\i) edge (l\i);
	    \node[main, fill=white!100] (k\i) at (\i,3.1) {LSTM};
        \path (i\i) edge [bend left] (k\i);
        \path (h\i) edge [bend left] (k\i);
	}
    \node (l4) at (4.5,2.3) {\ldots};
    \node (k4) at (4,3.1) {\ldots};
    \node (i4) at (4.5,.8) {\ldots};
    \node (h4) at (4,0) {\ldots};
    \node (x4) at (4,-1.3) {\ldots};
	\foreach \current/\next in {1/3,3/4,4/5,5/7} {
        \path (i\next) edge (i\current);
        \path (h\current) edge (h\next);
        \path (k\next) edge (k\current);
        \path (l\current) edge (l\next);
	}
    \node[main, fill=white!100] (mlp) at (4,4.6) {MLP};
	\foreach \i in {1,3} {
        \path (l\i) edge (mlp);
        \path (k\i) edge (mlp);
    }
    \coordinate (state) at (6.5,6.5);
    \path (state) edge [bend left] (mlp);
    \node (transition) at (4,5.8) {\textsc{Node}\textsubscript{U}};
    \path (mlp) edge (transition);
	\end{tikzpicture}
	\caption{Illustration of the \parser{} model.
		Top: parser state (stack, buffer and intermediate graph).
		Bottom: \parser{BiLTSM} architecture.
		Vector representation for the input tokens is computed
		by two layers of bidirectional LSTMs.
		The vectors for specific tokens are concatenated with
		embedding and numeric features from the parser state
		(for existing edge labels, number of children, etc.),
		and fed into the MLP for selecting the next transition.}
	\label{fig:model}
\end{figure}

\paragraph{Features.}
\parser{Sparse} uses binary indicator features representing
the words, POS tags, syntactic dependency labels and
existing edge labels related to the top four stack elements and the 
next three buffer elements, in addition to their children and grandchildren in the graph.
We also use bi- and trigram features based on these values \cite{zhang2009transition,zhu2013fast},
features related to discontinuous nodes
\cite[including separating punctuation and gap type]{maier2015discontinuous},
features representing existing edges and the number of parents and children,
as well as the past actions taken by the parser.
In addition, we use use a novel, UCCA-specific feature:
number of remote children.\footnote{See
Appendix~\ref{appendix:features} for a full list of used feature templates.}

For \parser{MLP} and \parser{BiLSTM},
we replace all indicator features by a
concatenation of the vector embeddings of all represented elements:
words, POS tags, syntactic dependency labels, edge labels, punctuation, gap type and parser actions.
These embeddings are initialized randomly.
We additionally use external word embeddings initialized with
pre-trained word2vec vectors \cite{mikolov2013efficient},\footnote{\url{
https://goo.gl/6ovEhC}} updated during training.
In addition to dropout between NN layers, we apply word dropout 
\cite{kiperwasser2016simple}: with a certain probability, the embedding for a
word is replaced with a zero vector. We do not apply word dropout to the external
word embeddings.

Finally, for all classifiers we add a novel real-valued feature to the input vector,
\textbf{ratio}, corresponding to the ratio between the number of terminals to number of nodes
in the graph $G$.
This feature serves as a regularizer for the creation of new nodes,
and should be beneficial for other transition-based constituency parsers too.

\paragraph{Training.}
For training the transition classifiers, we use a dynamic oracle \cite{goldberg2012dynamic},
i.e., an oracle that outputs a set of optimal transitions: when
applied to the current parser state, the gold
standard graph is reachable from the resulting state.
For example, the oracle would predict a \textsc{Node} transition if the stack 
has on its top a parent in the gold graph that has not been created,
but would predict a \textsc{Right-Edge} transition if the second stack
element is a parent of the
first element according to the gold graph and the edge between them has not been created.
The transition predicted by the classifier is deemed correct
and is applied to the parser state to reach the subsequent state,
if the transition is included in the set of optimal transitions.
Otherwise, a random optimal transition is applied,
and for the perceptron-based parser, the classifier's weights are updated according
to the perceptron update rule.

POS tags and syntactic dependency labels are extracted using spaCy
\cite{honnibal-johnson:2015:EMNLP}.\footnote{\url{https://spacy.io}}
We use the categorical cross-entropy objective function and optimize the
NN classifiers with the Adam optimizer \cite{kingma2014adam}.

\section{Experimental Setup}\label{sec:exp_setup}

\paragraph{Data.}
We conduct our experiments on the UCCA Wikipedia corpus (henceforth, \textit{Wiki}),
and use the English part of the UCCA \textit{Twenty Thousand Leagues Under the Sea}
English-French parallel corpus (henceforth, \textit{20K Leagues}) as
out-of-domain data.\footnote{\mbox{\url{http://cs.huji.ac.il/~oabend/ucca.html}}}
\tabref{table:data} presents some statistics for the two corpora.
We use passages of indices up to 676
of the \textit{Wiki} corpus as our training set, passages 688--808 as development set,
and passages 942--1028 as in-domain test set.
While UCCA edges can cross sentence boundaries, we adhere to the common
practice in semantic parsing and train our parsers on individual sentences,
discarding inter-relations between them (0.18\% of the edges).
We also discard linkage nodes and edges (as they often express inter-sentence
relations and are thus mostly redundant when applied at the sentence level)
as well as implicit nodes.\footnote{Appendix~\ref{appendix:extended_ucca}
further discusses linkage and implicit units.}
In the out-of-domain experiments, we apply the same parsers
(trained on the \textit{Wiki} training set) to the \textit{20K Leagues} corpus
without parameter re-tuning.

\begin{table}
	\scalebox{.9}{
	\begin{tabular}{l|ccc|c}
		& \multicolumn{3}{c|}{Wiki} & 20K \\
		& \small Train & \small Dev & \small Test & Leagues \\
		\hline
		\# passages & 300 & 34 & 33 & 154 \\
		\# sentences & 4268 & 454 & 503 & 506 \\
		\hline
		\# nodes & 298,993 & 33,704 & 35,718 & 29,315 \\
		\% terminal & 42.96 & 43.54 & 42.87 & 42.09 \\
		\% non-term. & 58.33 & 57.60 & 58.35 & 60.01 \\
		\% discont. & 0.54 & 0.53 & 0.44 & 0.81 \\
		\% reentrant & 2.38 & 1.88 & 2.15 & 2.03 \\
		\hline
		\# edges & 287,914 & 32,460 & 34,336 & 27,749 \\
		\% primary & 98.25 & 98.75 & 98.74 & 97.73 \\
		\% remote & 1.75 & 1.25 & 1.26 & 2.27 \\
		\hline
		\multicolumn{3}{l}{\footnotesize Average per non-terminal node} \\
		\# children & 1.67 & 1.68 & 1.66 & 1.61 
	\end{tabular}
	}
	\caption{Statistics of the \textit{Wiki} and \textit{20K Leagues} UCCA corpora.
		All counts exclude the root node, implicit nodes, and linkage nodes and edges.
	}
	\label{table:data}
\end{table}

\paragraph{Implementation.}
We use the DyNet package \cite{neubig2017dynet} for implementing the NN classifiers.
Unless otherwise noted, we use the default values provided by the package.
See Appendix~\ref{appendix:hyperparameters} for the hyperparameter values we found by tuning
on the development set.

\paragraph{Evaluation.}
We define a simple measure for comparing UCCA structures
$G_p=(V_p,E_p,\ell_p)$ and $G_g=(V_g,E_g,\ell_g)$,
the predicted and gold-standard graphs, respectively, over the same
sequence of terminals $W = \{w_1,\ldots,w_n\}$.
For an edge $e=(u,v)$ in either graph,
$u$ being the parent and $v$ the child, its yield $y(e) \subseteq W$ is the
set of terminals in $W$ that are descendants of $v$.
Define the set of \textit{mutual edges} between $G_p$ and $G_g$:

\vspace{-.6cm}

{\small
\begin{multline*}
    M(G_p,G_g) = \\
    \left\{(e_1,e_2) \in E_p \times E_g \;|\;
    y(e_1) = y(e_2) \wedge \ell_p(e_1)=\ell_g(e_2)\right\}
\end{multline*}
}

\vspace{-.6cm}

Labeled precision and recall are defined by dividing $|M(G_p,G_g)|$ by $|E_p|$ and $|E_g|$, respectively,
and F-score by taking their harmonic mean.
We report two variants of this measure: one where we consider only primary edges,
and another for remote edges (see \secref{sec:ucca}).
Performance on remote edges is of pivotal importance in this investigation,
which focuses on extending the class of graphs supported by statistical parsers.

We note that the measure collapses to the standard
PARSEVAL constituency evaluation measure if $G_p$ and $G_g$ are trees.
Punctuation is excluded from the evaluation, but not from the datasets.

\begin{figure}
	\centering
	\scalebox{.9}{
	\begin{dependency}[theme = simple]
	\begin{deptext}[column sep=.7em,ampersand replacement=\^]
	After \^ graduation \^ , \^ John \^ moved \^ to \^ Paris \\
	\end{deptext}
		\depedge{2}{1}{L}
		\depedge[edge start x offset=7pt]{2}{3}{U}
		\depedge[edge start x offset=3pt,dashed]{2}{4}{A}
		\depedge{5}{4}{A}
		\depedge{2}{5}{H}
		\depedge{7}{6}{R}
		\depedge{5}{7}{A}
	\end{dependency}
	}
	\scalebox{.9}{
	\begin{dependency}[theme = simple]
	\begin{deptext}[column sep=.7em,ampersand replacement=\^]
	John \^ gave \^ everything \^ up \\
	\end{deptext}
		\depedge{2}{1}{A}
		\depedge{2}{3}{A}
		\depedge{2}{4}{C}
	\end{dependency}
	}
	\scalebox{.9}{
	\begin{dependency}[theme = simple]
	\begin{deptext}[column sep=.7em,ampersand replacement=\^]
	John \^ and \^ Mary \^ went \^ home \\
	\end{deptext}
		\depedge{4}{1}{A}
		\depedge[edge start x offset=3pt]{1}{2}{N}
		\depedge{1}{3}{C}
		\depedge{4}{5}{A}
	\end{dependency}
	}
	\caption{Bilexical graph approximation (dependency graph) for the sentences in \figref{fig:examples}.}
	\label{fig:bilexical_example}
\end{figure}

\begin{table*}
	\begin{tabular}{l|ccc|ccc||ccc|ccc}
		& \multicolumn{6}{c||}{Wiki (in-domain)} & \multicolumn{6}{c}{20K Leagues (out-of-domain)} \\
		& \multicolumn{3}{c|}{Primary} & \multicolumn{3}{c||}{Remote}
		& \multicolumn{3}{c|}{Primary} & \multicolumn{3}{c}{Remote} \\
		& \textbf{LP} & \textbf{LR} & \textbf{LF} & \textbf{LP} & \textbf{LR} & \textbf{LF}
		& \textbf{LP} & \textbf{LR} & \textbf{LF} & \textbf{LP} & \textbf{LR} & \textbf{LF} \\
		\hline
		\parser{Sparse}
		& 64.5 & 63.7 & 64.1 & 19.8 & 13.4 & 16
		& 59.6 & 59.9 & 59.8 & 22.2 & 7.7 & 11.5 \\
		\parser{MLP}
		& 65.2 & 64.6 & 64.9 & 23.7 & 13.2 & 16.9
		& 62.3 & 62.6 & 62.5 & 20.9 & 6.3 & 9.7 \\
		\parser{BiLSTM}
		& 74.4 & 72.7 & \textbf{73.5} & 47.4 & 51.6 & \textbf{49.4}
		& 68.7 & 68.5 & \textbf{68.6} & 38.6 & 18.8 & \textbf{25.3} \\
		\hline
		\multicolumn{8}{l}{\rule{0pt}{2ex} \footnotesize
		Bilexical Approximation (Dependency DAG Parsers)} \\
		\small Upper Bound
		& & & \small 91 & & & \small 58.3
		& & & \small 91.3 & & & \small 43.4 \\
		DAGParser
		& 61.8 & 55.8 & 58.6 & 9.5 & 0.5 & 1
		& 56.4 & 50.6 & 53.4 & -- & 0 & 0 \\
		TurboParser
		& 57.7 & 46 & 51.2 & 77.8 & 1.8 & 3.7
		& 50.3 & 37.7 & 43.1 & 100 & 0.4 & 0.8 \\
		\hline
		\multicolumn{8}{l}{\rule{0pt}{2ex} \footnotesize
		Tree Approximation (Constituency Tree Parser)} \\
		\small Upper Bound
		& & & \small 100 & & & \small --
		& & & \small 100 & & & \small -- \\
		\textsc{uparse}
		& 60.9 & 61.2 & 61.1 & -- & -- & --
		& 52.7 & 52.8 & 52.8 & -- & -- & -- \\
		\hline
		\multicolumn{8}{l}{\rule{0pt}{2ex} \footnotesize
		Bilexical Tree Approximation (Dependency Tree Parsers)} \\
		\small Upper Bound
		& & & \small 91 & & & \small --
		& & & \small 91.3 & & & \small -- \\
		MaltParser
		& 62.8 & 57.7 & 60.2 & -- & -- & --
		& 57.8 & 53 & 55.3 & -- & -- & -- \\
		LSTM Parser
		& 73.2 & 66.9 & 69.9 & -- & -- & --
		& 66.1 & 61.1 & 63.5 & -- & -- & --
	\end{tabular}
	\caption{
	  Experimental results, in percents, on the \textit{Wiki} test set (left)
	  and the \textit{20K Leagues} set (right).
	  Columns correspond to labeled precision, recall and F-score,
	  for both primary and remote edges.
	  F-score upper bounds are reported for the conversions.
	  For the tree approximation experiments, only primary edges scores are reported,
	  as they are unable to predict remote edges.
	  \parser{BiLSTM} obtains the highest F-scores in all metrics, surpassing the
	  bilexical parsers, tree parsers and other classifiers.
	}
	\label{table:results}
\end{table*}

\paragraph{Comparison to bilexical graph parsers.}
As no direct comparison with existing parsers is possible,
we compare \parser{} to bilexical dependency graph parsers,
which support reentrancy and discontinuity but not non-terminal nodes.

To facilitate the comparison, we convert our training set into bilexical graphs
(see examples in \figref{fig:bilexical_example}),
train each of the parsers, and evaluate them by applying them to the test set
and then reconstructing UCCA graphs, which are compared with the gold standard.
The conversion to bilexical graphs is done by heuristically selecting a head terminal for each
non-terminal node, and attaching all terminal descendents to the head terminal.
In the inverse conversion, we traverse the bilexical graph in topological order,
creating non-terminal parents for all terminals, and attaching them to the previously-created
non-terminals corresponding to the bilexical
heads.\footnote{See Appendix~\ref{appendix:conversion} for a detailed description of
the conversion procedures.}

In \secref{sec:results} we report the upper bounds on the achievable scores due to the
error resulting from the removal of non-terminal nodes.

\begin{figure}
	\centering
	\scalebox{.9}{
	  \begin{tikzpicture}[level distance=10mm, ->]
	    \node (ROOT) [fill=black, circle] {}
	      child {node (After) {After} edge from parent node[left] {\scriptsize $L$}}
	      child {node (graduation) [fill=black, circle] {}
	      {
	        child {node {graduation} edge from parent node[left] {\scriptsize $P$}}
	      } edge from parent node[left] {\scriptsize $H$} }
	      child {node {,} edge from parent node[right] {\scriptsize $U$}}
	      child {node (moved) [fill=black, circle] {}
	      {
	        child {node (John) {John} edge from parent node[left] {\scriptsize $A$}}
	        child {node {moved} edge from parent node[left] {\scriptsize $P$}}
	        child {node [fill=black, circle] {}
	        {
	          child {node {to} edge from parent node[left] {\scriptsize $R$}}
	          child {node {Paris} edge from parent node[left] {\scriptsize $C$}}
	        } edge from parent node[left] {\scriptsize $A$} }
	      } edge from parent node[right] {\scriptsize $H$} }
	      ;
	  \end{tikzpicture}
	}
	\scalebox{.9}{
	\begin{dependency}[theme = simple]
	\begin{deptext}[column sep=.7em,ampersand replacement=\^]
	After \^ graduation \^ , \^ John \^ moved \^ to \^ Paris \\
	\end{deptext}
		\depedge{2}{1}{L}
		\depedge{2}{3}{U}
		\depedge{5}{4}{A}
		\depedge{2}{5}{H}
		\depedge{7}{6}{R}
		\depedge{5}{7}{A}
	\end{dependency}
  }
  \caption{Tree approximation (constituency) for the sentence in \figref{fig:graduation} (top),
  and bilexical tree approximation (dependency) for the same sentence (bottom).
  These are identical to the original graphs,
  apart from the removal of remote edges.}
  \label{fig:tree_example}
\end{figure}
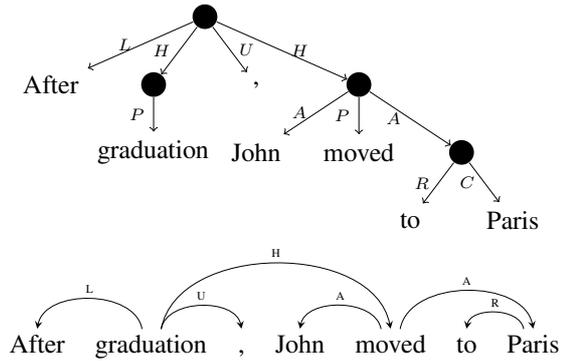

\paragraph{Comparison to tree parsers.}
For completeness,
and as parsing technology is considerably more mature for tree (rather than graph) parsing,
we also perform a \textit{tree approximation} experiment, converting UCCA to (bilexical) trees
and evaluating constituency and dependency tree parsers on them
(see examples in \figref{fig:tree_example}).
Our approach is similar
to the tree approximation approach used for dependency graph parsing
\cite{agic2015semantic,fernandez2015parsing},
where dependency graphs were converted into dependency trees
and then parsed by dependency tree parsers.
In our setting, the conversion to trees consists simply of removing remote edges from the 
graph, and then to bilexical trees by applying the same procedure as for bilexical graphs.

\paragraph{Baseline parsers.}
We evaluate two bilexical graph semantic dependency parsers:
DAGParser \cite{ribeyre-villemontedelaclergerie-seddah:2014:SemEval}, the leading 
transition-based parser in SemEval 2014 \cite{oepen2014semeval} and
TurboParser \cite{almeida-martins:2015:SemEval},
a graph-based parser from SemEval 2015 
\cite{oepen2015semeval};
\textsc{uparse} \cite{maier-lichte:2016:DiscoNLP},
a transition-based constituency parser supporting discontinuous constituents;
and two bilexical tree parsers:
MaltParser \cite{nivre2007maltparser},
and the stack LSTM-based parser of
\citet[henceforce ``LSTM Parser'']{dyer2015transition}.
Default settings are used in all cases.\footnote{For
MaltParser we use the \textsc{ArcEager} transition set and SVM classifier.
Other configurations yielded lower scores.}
DAGParser and \textsc{uparse} use beam search by default, with a beam size of 5 and 4
respectively. The other parsers are greedy.

\section{Results}\label{sec:results}

\tabref{table:results} presents our main experimental results, as well as
upper bounds for the baseline parsers,
reflecting the error resulting from the conversion.\footnote{The low
upper bound for remote edges is partly due to the removal of implicit nodes (not supported
in bilexical representations), where the whole sub-graph headed by such nodes,
often containing remote edges, must be discarded.}

DAGParser and \textsc{uparse} are most directly comparable to
\parser{Sparse}, as they also use a perceptron classifier with sparse features.
\parser{Sparse} considerably outperforms both, where
DAGParser does not predict any remote edges in the out-of-domain setting.
TurboParser fares worse in this comparison, despite somewhat better results on
remote edges.
The LSTM parser of \citet{dyer2015transition} obtains the highest primary F-score
among the baseline parsers, with a considerable margin.

Using a feedforward NN and embedding features,
\parser{MLP} obtains higher scores than \parser{Sparse},
but is outperformed by the LSTM parser on primary edges.
However, using better input encoding
allowing virtual look-ahead and look-behind in the token representation,
\parser{BiLSTM} obtains substantially higher scores than \parser{MLP}
and all other parsers, on both primary and remote edges,
both in the in-domain and out-of-domain settings.
Its performance in absolute terms, of 73.5\% F-score on primary edges,
is encouraging in light of
UCCA's inter-annotator agreement of 80--85\%
F-score on them \cite{abend2013universal}.

The parsers resulting from tree approximation are unable to recover any remote edges,
as these are removed in the conversion.\footnote{We
also experimented with a simpler version of \parser{} lacking
\textsc{Remote} transitions, obtaining an increase of up to 2 labeled F-score
points on primary edges, at the cost of not being able to predict remote edges.}
The bilexical DAG parsers are quite limited in this respect as well.
While some of the DAG parsers' difficulty can be attributed to the conversion upper bound of 58.3\%,
this in itself cannot account for their
poor performance on remote edges, which is an order of magnitude lower than that
of \parser{BiLSTM}.


\section{Related Work}\label{sec:related_work}

While earlier work on anchored\footnote{By {\it anchored} we mean that the semantic representation
directly corresponds to the words and phrases of the text.}
semantic parsing has mostly concentrated on shallow semantic analysis,
focusing on semantic role labeling of verbal argument structures,
the focus has recently shifted to parsing of more elaborate representations that account
for a wider range of phenomena \cite{abend2017the}.

\paragraph{Grammar-Based Parsing.}
Linguistically expressive grammars such as HPSG \cite{PandS:94}, CCG \cite{Steedman:00} and TAG \cite{Joshi:97}
provide a theory of the syntax-semantics interface, and have been used as a basis for semantic parsers
by defining compositional semantics on top of them \cite[among others]{Flic:00,bos2005towards}.
Depending on the grammar and the implementation, such semantic parsers can support
some or all of the structural properties UCCA exhibits.
Nevertheless, this line of work differs from our approach in two important ways.
First, the \textit{representations} are different. UCCA does not attempt to model
the syntax-semantics interface and is thus less coupled with syntax.
Second, while grammar-based \textit{parsers} explicitly model syntax,
our approach directly models the relation between tokens and semantic structures,
without explicit composition rules.

\paragraph{Broad-Coverage Semantic Parsing.}
Most closely related to this work is Broad-Coverage Semantic Dependency Parsing (SDP),
addressed in two SemEval tasks \cite{oepen2014semeval,oepen2015semeval}.
Like UCCA parsing, SDP addresses a wide range of semantic phenomena,
and supports discontinuous units and reentrancy.
In SDP, however, bilexical dependencies are used,
and a head must be selected for every relation---even in constructions that have no clear head,
such as coordination \cite{Ivanova2012who}.
The use of non-terminal nodes is a simple way to avoid this liability.
SDP also differs from UCCA in the type of distinctions it makes,
which are more tightly coupled with syntactic considerations,
where UCCA aims to capture purely semantic cross-linguistically applicable notions.
For instance, the ``poss'' label in the DM target representation is used to
annotate syntactic possessive constructions, regardless of whether they correspond to
semantic ownership (e.g., ``John's dog'') or other semantic relations,
such as marking an argument of a nominal predicate (e.g., ``John's kick'').
UCCA reflects the difference between these constructions.

Recent interest in SDP has yielded numerous works on graph parsing
\cite{ribeyre-villemontedelaclergerie-seddah:2014:SemEval,thomson-EtAl:2014:SemEval,almeida-martins:2015:SemEval,du-EtAl:2015:SemEval}, including
tree approximation \cite{agic-koller:2014:SemEval,schluter-EtAl:2014:SemEval}
and joint syntactic/semantic parsing
\cite{henderson2013multilingual,swayamdipta-EtAl:2016:CoNLL}.

\paragraph{Abstract Meaning Representation.}
Another line of work addresses parsing into AMRs
\cite{flanigan2014discriminative,vanderwende2015amr,pust2015parsing,artzi2015broad},
which, like UCCA, abstract away from syntactic distinctions
and represent meaning directly, using OntoNotes predicates \cite{weischedel2013ontonotes}.
Events in AMR may also be evoked by non-verbal predicates, including possessive constructions.

Unlike in UCCA, the alignment between AMR concepts and the text is not explicitly marked.
While sharing much of this work's motivation, not anchoring the representation in the text
complicates the parsing task, as it requires
the alignment to be automatically (and imprecisely) detected.
Indeed, despite considerable technical effort
\cite{flanigan2014discriminative,pourdamghani2014aligning,werling2015robust},
concept identification is only about 80\%--90\% accurate.
Furthermore, anchoring allows breaking down sentences into semantically meaningful sub-spans,
which is useful for many applications \cite{fernandez2015parsing,birch2016hume}.

Several transition-based AMR parsers have been proposed:
CAMR assumes syntactically parsed input,
processing dependency trees into AMR
\cite{wang-xue-pradhan:2015:ACL-IJCNLP,wang2015transition,wang-EtAl:2016:SemEval,goodman2016noise}.
In contrast, the parsers of \citet{damonte-17} and \citet{zhou2016amr}
do not require syntactic pre-processing.
\citet{damonte-17} perform concept identification using
a simple heuristic selecting the most frequent graph for each token, and
\citet{zhou2016amr} perform concept identification and parsing jointly.
UCCA parsing does not require separately aligning the input tokens to the graph.
\parser{} creates non-terminal units as part of the parsing process.

Furthermore, existing transition-based AMR parsers are not general DAG parsers.
They are only able to predict a subset of reentrancies and discontinuities,
as they may remove nodes before their parents have been predicted
\cite{damonte-17}.
They are thus limited to a sub-class of AMRs in particular,
and specifically cannot produce arbitrary DAG parses.
\parser{}'s transition set, on the other hand, allows general DAG
parsing.\footnote{See Appendix~\ref{appendix:completeness_proof} for a proof sketch for the
completeness of \parser{}'s transition set.}

\section{Conclusion}\label{sec:conclusion}

We present \parser{}, the first parser for UCCA.
Evaluated in in-domain and out-of-domain settings, we show that coupled with a
NN classifier and BiLSTM feature extractor,
it accurately predicts UCCA graphs from text, outperforming a variety of
strong baselines by a margin.

Despite the recent diversity of semantic parsing work,
the effectiveness of different approaches for
structurally and semantically different schemes is not well-understood
\cite{kuhlmann2016towards}.
Our contribution to this literature is a general parser
that supports multiple parents, discontinuous units and non-terminal nodes.

Future work will evaluate \parser{} in a multilingual setting,
assessing UCCA's cross-linguistic applicability.
We will also apply the \parser{} transition scheme to different target
representations, including AMR and SDP, exploring the limits of its generality.
In addition, we will explore different conversion procedures \cite{kong-15}
to compare different representations,
suggesting ways for a data-driven design of semantic annotation.

A parser for UCCA will enable using the framework for new tasks,
in addition to existing applications such as machine translation
evaluation \cite{birch2016hume}.
We believe UCCA's merits in providing a cross-linguistically applicable,
broad-coverage annotation will support ongoing efforts to incorporate deeper
semantic structures into various applications,
such as sentence simplification \cite{narayan2014hybrid} and summarization \cite{liu2015toward}.

\section*{Acknowledgments}

This work was supported by the HUJI Cyber Security Research Center
in conjunction with the Israel National Cyber Bureau in the Prime Minister's Office,
and by the Intel Collaborative Research Institute for Computational Intelligence (ICRI-CI).
The first author was supported by a fellowship from the
Edmond and Lily Safra Center for Brain Sciences.
We thank Wolfgang Maier, Nathan Schneider, Elior Sulem
and the anonymous reviewers for their helpful comments.

\bibliography{references}

\begin{thebibliography}{}
\expandafter\ifx\csname natexlab\endcsname\relax\def\natexlab#1{#1}\fi

\bibitem[{Abend and Rappoport(2013)}]{abend2013universal}
Omri Abend and Ari Rappoport. 2013.
\newblock {U}niversal {C}onceptual {C}ognitive {A}nnotation ({UCCA}).
\newblock In {\em Proc. of ACL\/}. pages 228--238.

\bibitem[{Abend and Rappoport(2017)}]{abend2017the}
Omri Abend and Ari Rappoport. 2017.
\newblock The state of the art in semantic representation.
\newblock In {\em Proc. of ACL\/}.
\newblock To appear.

\bibitem[{Abend et~al.(2017)Abend, Yerushalmi, and
  Rappoport}]{abend2017uccaapp}
Omri Abend, Shai Yerushalmi, and Ari Rappoport. 2017.
\newblock {UCCAApp}: Web-application for syntactic and semantic phrase-based
  annotation.
\newblock In {\em Proc. of ACL: System Demonstration Papers\/}.
\newblock To appear.

\bibitem[{Agi\'{c} and Koller(2014)}]{agic-koller:2014:SemEval}
\v{Z}eljko Agi\'{c} and Alexander Koller. 2014.
\newblock Potsdam: Semantic dependency parsing by bidirectional graph-tree
  transformations and syntactic parsing.
\newblock In {\em Proc. of SemEval\/}. pages 465--470.

\bibitem[{Agi{\'c} et~al.(2015)Agi{\'c}, Koller, and Oepen}]{agic2015semantic}
{\v{Z}}eljko Agi{\'c}, Alexander Koller, and Stephan Oepen. 2015.
\newblock Semantic dependency graph parsing using tree approximations.
\newblock In {\em Proc. of IWCS\/}. pages 217--227.

\bibitem[{Almeida and Martins(2015)}]{almeida-martins:2015:SemEval}
Mariana S.~C. Almeida and Andr\'{e} F.~T. Martins. 2015.
\newblock Lisbon: Evaluating {T}urbo{S}emantic{P}arser on multiple languages
  and out-of-domain data.
\newblock In {\em Proc. of SemEval\/}. pages 970--973.

\bibitem[{Ambati et~al.(2015)Ambati, Deoskar, Johnson, and
  Steedman}]{ambati2015incremental}
Bharat~Ram Ambati, Tejaswini Deoskar, Mark Johnson, and Mark Steedman. 2015.
\newblock An incremental algorithm for transition-based {CCG} parsing.
\newblock In {\em Proc. of NAACL\/}. pages 53--63.

\bibitem[{Ambati et~al.(2016)Ambati, Deoskar, and
  Steedman}]{ambati-deoskar-steedman:2016:N16-1}
Bharat~Ram Ambati, Tejaswini Deoskar, and Mark Steedman. 2016.
\newblock Shift-reduce {CCG} parsing using neural network models.
\newblock In {\em Proc. of NAACL-HLT\/}. pages 447--453.

\bibitem[{Andor et~al.(2016)Andor, Alberti, Weiss, Severyn, Presta, Ganchev,
  Petrov, and Collins}]{andor2016globally}
Daniel Andor, Chris Alberti, David Weiss, Aliaksei Severyn, Alessandro Presta,
  Kuzman Ganchev, Slav Petrov, and Michael Collins. 2016.
\newblock Globally normalized transition-based neural networks.
\newblock In {\em Proc. of ACL\/}. pages 2442--2452.

\bibitem[{Artzi et~al.(2015)Artzi, Lee, and Zettlemoyer}]{artzi2015broad}
Yoav Artzi, Kenton Lee, and Luke Zettlemoyer. 2015.
\newblock Broad-coverage {CCG} semantic parsing with {AMR}.
\newblock In {\em Proc. of EMNLP\/}. pages 1699--1710.

\bibitem[{Banarescu et~al.(2013)Banarescu, Bonial, Cai, Georgescu, Griffitt,
  Hermjakob, Knight, Palmer, and Schneider}]{banarescu2013abstract}
Laura Banarescu, Claire Bonial, Shu Cai, Madalina Georgescu, Kira Griffitt, Ulf
  Hermjakob, Kevin Knight, Martha Palmer, and Nathan Schneider. 2013.
\newblock {A}bstract {M}eaning {R}epresentation for sembanking.
\newblock In {\em Proc. of the Linguistic Annotation Workshop\/}.

\bibitem[{Birch et~al.(2016)Birch, Abend, Bojar, and Haddow}]{birch2016hume}
Alexandra Birch, Omri Abend, Ond\v{r}ej Bojar, and Barry Haddow. 2016.
\newblock {HUME}: Human {UCCA}-based evaluation of machine translation.
\newblock In {\em Proc. of EMNLP\/}. pages 1264--1274.

\bibitem[{Bos(2005)}]{bos2005towards}
Johan Bos. 2005.
\newblock Towards wide-coverage semantic interpretation.
\newblock In {\em Proc. of IWCS\/}. volume~6, pages 42--53.

\bibitem[{Chen and Manning(2014)}]{chen2014fast}
Danqi Chen and Christopher Manning. 2014.
\newblock A fast and accurate dependency parser using neural networks.
\newblock In {\em Proc. of EMNLP\/}. pages 740--750.

\bibitem[{Collins(1997)}]{Coll:97}
Michael Collins. 1997.
\newblock Three generative lexicalized models for statistical parsing.
\newblock In {\em Proc. of ACL\/}. ACL, Madrid, pages 16--23.

\bibitem[{Collins and Roark(2004)}]{Coll:04}
Michael Collins and Brian Roark. 2004.
\newblock Incremental parsing with the perceptron algorithm.
\newblock In {\em Proc. of ACL\/}. pages 111--118.

\bibitem[{Croft and Cruse(2004)}]{croft2004cognitive}
William Croft and D~Alan Cruse. 2004.
\newblock {\em Cognitive linguistics\/}.
\newblock Cambridge University Press.

\bibitem[{Damonte et~al.(2017)Damonte, Cohen, and Satta}]{damonte-17}
Marco Damonte, Shay~B. Cohen, and Giorgio Satta. 2017.
\newblock An incremental parser for abstract meaning representation.
\newblock In {\em Proceedings of {EACL}\/}.

\bibitem[{Dixon(2010{\natexlab{a}})}]{Dixon:10b}
Robert M.~W. Dixon. 2010{\natexlab{a}}.
\newblock {\em Basic Linguistic Theory: Grammatical Topics\/}, volume~2.
\newblock Oxford University Press.

\bibitem[{Dixon(2010{\natexlab{b}})}]{Dixon:10a}
Robert M.~W. Dixon. 2010{\natexlab{b}}.
\newblock {\em Basic Linguistic Theory: Methodology\/}, volume~1.
\newblock Oxford University Press.

\bibitem[{Dixon(2012)}]{Dixon:12}
Robert M.~W. Dixon. 2012.
\newblock {\em Basic Linguistic Theory: Further Grammatical Topics\/},
  volume~3.
\newblock Oxford University Press.

\bibitem[{Du et~al.(2015)Du, Zhang, Zhang, Sun, and Wan}]{du-EtAl:2015:SemEval}
Yantao Du, Fan Zhang, Xun Zhang, Weiwei Sun, and Xiaojun Wan. 2015.
\newblock Peking: Building semantic dependency graphs with a hybrid parser.
\newblock In {\em Proc. of SemEval\/}. pages 927--931.

\bibitem[{Dyer et~al.(2015)Dyer, Ballesteros, Ling, Matthews, and
  Smith}]{dyer2015transition}
Chris Dyer, Miguel Ballesteros, Wang Ling, Austin Matthews, and Noah~A. Smith.
  2015.
\newblock Transition-based dependeny parsing with stack long short-term memory.
\newblock In {\em Proc. of ACL\/}. pages 334--343.

\bibitem[{Fern{\'a}ndez-Gonz{\'a}lez and Martins(2015)}]{fernandez2015parsing}
Daniel Fern{\'a}ndez-Gonz{\'a}lez and Andr{\'e}~FT Martins. 2015.
\newblock Parsing as reduction.
\newblock In {\em Proc. of ACL\/}. pages 1523--1533.

\bibitem[{Flanigan et~al.(2014)Flanigan, Thomson, Carbonell, Dyer, and
  Smith}]{flanigan2014discriminative}
Jeffrey Flanigan, Sam Thomson, Jaime Carbonell, Chris Dyer, and Noah~A. Smith.
  2014.
\newblock A discriminative graph-based parser for the abstract meaning
  representation.
\newblock In {\em Proc. of ACL\/}. pages 1426--1436.

\bibitem[{Flickinger(2000)}]{Flic:00}
Daniel Flickinger. 2000.
\newblock On building a more efficient grammar by exploiting types.
\newblock In {\em Collaborative Language Engineering\/}, CLSI, Stanford,~CA,
  volume~6, pages 15--28.

\bibitem[{Goldberg and Elhadad(2011)}]{goldberg2011learning}
Yoav Goldberg and Michael Elhadad. 2011.
\newblock Learning sparser perceptron models.
\newblock Technical report.

\bibitem[{Goldberg and Nivre(2012)}]{goldberg2012dynamic}
Yoav Goldberg and Joakim Nivre. 2012.
\newblock A dynamic oracle for arc-eager dependency parsing.
\newblock In {\em Proc. of COLING\/}. pages 959--976.

\bibitem[{Goodman et~al.(2016)Goodman, Vlachos, and
  Naradowsky}]{goodman2016noise}
James Goodman, Andreas Vlachos, and Jason Naradowsky. 2016.
\newblock Noise reduction and targeted exploration in imitation learning for
  {A}bstract {M}eaning {R}epresentation parsing.
\newblock In {\em Proc. of ACL\/}. pages 1--11.

\bibitem[{Henderson et~al.(2013)Henderson, Merlo, Titov, and
  Musillo}]{henderson2013multilingual}
James Henderson, Paola Merlo, Ivan Titov, and Gabriele Musillo. 2013.
\newblock Multilingual joint parsing of syntactic and semantic dependencies
  with a latent variable model.
\newblock {\em Computational Linguistics\/} 39(4):949--998.

\bibitem[{Honnibal and Johnson(2015)}]{honnibal-johnson:2015:EMNLP}
Matthew Honnibal and Mark Johnson. 2015.
\newblock An improved non-monotonic transition system for dependency parsing.
\newblock In {\em Proc. of EMNLP\/}. pages 1373--1378.

\bibitem[{Ivanova et~al.(2012)Ivanova, Oepen, {\O}vrelid, and
  Flickinger}]{Ivanova2012who}
Angelina Ivanova, Stephan Oepen, Lilja {\O}vrelid, and Dan Flickinger. 2012.
\newblock Who did what to whom? {A} contrastive study of syntacto-semantic
  dependencies.
\newblock In {\em Proc. of LAW\/}. pages 2--11.

\bibitem[{Joshi and Schabes(1997)}]{Joshi:97}
Aravind Joshi and Yves Schabes. 1997.
\newblock {T}ree-{A}djoining {G}rammars.
\newblock In Grzegorz Rozenberg and Arto Salomaa, editors, {\em Handbook of
  Formal Languages\/}, Springer, Berlin, volume~3, pages 69--124.

\bibitem[{Kingma and Ba(2014)}]{kingma2014adam}
Diederik~P. Kingma and Jimmy Ba. 2014.
\newblock Adam: {A} method for stochastic optimization.
\newblock {\em CoRR\/} abs/1412.6980.

\bibitem[{Kiperwasser and Goldberg(2016)}]{kiperwasser2016simple}
Eliyahu Kiperwasser and Yoav Goldberg. 2016.
\newblock Simple and accurate dependency parsing using bidirectional {LSTM}
  feature representations.
\newblock {\em TACL\/} 4:313--327.

\bibitem[{Kong et~al.(2015)Kong, Rush, and Smith}]{kong-15}
Lingpeng Kong, Alexander~M. Rush, and Noah~A. Smith. 2015.
\newblock Transforming dependencies into phrase structures.
\newblock In {\em Proc. of NAACL HLT\/}.

\bibitem[{Kuhlmann and Oepen(2016)}]{kuhlmann2016towards}
Marco Kuhlmann and Stephan Oepen. 2016.
\newblock Towards a catalogue of linguistic graph banks.
\newblock {\em Computational Linguistics\/} .

\bibitem[{Liu et~al.(2015)Liu, Flanigan, Thomson, Sadeh, and
  Smith}]{liu2015toward}
Fei Liu, Jeffrey Flanigan, Sam Thomson, Norman Sadeh, and Noah~A. Smith. 2015.
\newblock Toward abstractive summarization using semantic representations.
\newblock In {\em Proc. of NAACL\/}. pages 1077--1086.

\bibitem[{Maier(2015)}]{maier2015discontinuous}
Wolfgang Maier. 2015.
\newblock Discontinuous incremental shift-reduce parsing.
\newblock In {\em Proc. of ACL\/}. pages 1202--1212.

\bibitem[{Maier and Lichte(2009)}]{Maier:Lichte:11}
Wolfgang Maier and Timm Lichte. 2009.
\newblock Characterizing discontinuity in constituent treebanks.
\newblock In {\em Proc. of Formal Grammar\/}. Springer, Bordeaux, France,
  number 5591 in Lecture Notes in Artificial Intelligence, pages 167--182.

\bibitem[{Maier and Lichte(2016)}]{maier-lichte:2016:DiscoNLP}
Wolfgang Maier and Timm Lichte. 2016.
\newblock Discontinuous parsing with continuous trees.
\newblock In {\em Proc. of Workshop on Discontinuous Structures in NLP\/}.
  pages 47--57.

\bibitem[{Mikolov et~al.(2013)Mikolov, Chen, Corrado, and
  Dean}]{mikolov2013efficient}
Tomas Mikolov, Kai Chen, Greg Corrado, and Jeffrey Dean. 2013.
\newblock Efficient estimation of word representations in vector space.
\newblock {\em CoRR\/} abs/1301.3781.

\bibitem[{Misra and Artzi(2016)}]{dipendra2016neural}
Dipendra~K Misra and Yoav Artzi. 2016.
\newblock Neural shift-reduce {CCG} semantic parsing.
\newblock In {\em Proc. of EMNLP\/}. pages 1775--1786.

\bibitem[{Narayan and Gardent(2014)}]{narayan2014hybrid}
Shashi Narayan and Claire Gardent. 2014.
\newblock Hybrid simplification using deep semantics and machine translation.
\newblock In {\em Proc. of ACL\/}. pages 435--445.

\bibitem[{Neubig et~al.(2017)Neubig, Dyer, Goldberg, Matthews, Ammar,
  Anastasopoulos, Ballesteros, Chiang, Clothiaux, Cohn, Duh, Faruqui, Gan,
  Garrette, Ji, Kong, Kuncoro, Kumar, Malaviya, Michel, Oda, Richardson,
  Saphra, Swayamdipta, and Yin}]{neubig2017dynet}
Graham Neubig, Chris Dyer, Yoav Goldberg, Austin Matthews, Waleed Ammar,
  Antonios Anastasopoulos, Miguel Ballesteros, David Chiang, Daniel Clothiaux,
  Trevor Cohn, Kevin Duh, Manaal Faruqui, Cynthia Gan, Dan Garrette, Yangfeng
  Ji, Lingpeng Kong, Adhiguna Kuncoro, Gaurav Kumar, Chaitanya Malaviya, Paul
  Michel, Yusuke Oda, Matthew Richardson, Naomi Saphra, Swabha Swayamdipta, and
  Pengcheng Yin. 2017.
\newblock Dy{N}et: The dynamic neural network toolkit.
\newblock {\em arXiv preprint arXiv:1701.03980\/} .

\bibitem[{Nivre(2003)}]{Nivre03anefficient}
Joakim Nivre. 2003.
\newblock An efficient algorithm for projective dependency parsing.
\newblock In {\em Proc. of IWPT\/}. pages 149--160.

\bibitem[{Nivre(2009)}]{nivre2009non}
Joakim Nivre. 2009.
\newblock Non-projective dependency parsing in expected linear time.
\newblock In {\em Proc. of ACL\/}. pages 351--359.

\bibitem[{Nivre et~al.(2007)Nivre, Hall, Nilsson, Chanev, Eryigit, K{\"u}bler,
  Marinov, and Marsi}]{nivre2007maltparser}
Joakim Nivre, Johan Hall, Jens Nilsson, Atanas Chanev, G{\"u}lsen Eryigit,
  Sandra K{\"u}bler, Svetoslav Marinov, and Erwin Marsi. 2007.
\newblock {M}alt{P}arser: A language-independent system for data-driven
  dependency parsing.
\newblock {\em Natural Language Engineering\/} 13(02):95--135.

\bibitem[{Oepen et~al.(2015)Oepen, Kuhlmann, Miyao, Zeman, Cinkov{\'a},
  Flickinger, Haji\v{c}, and Ure\v{s}ov{\'a}}]{oepen2015semeval}
Stephan Oepen, Marco Kuhlmann, Yusuke Miyao, Daniel Zeman, Silvie Cinkov{\'a},
  Dan Flickinger, Jan Haji\v{c}, and Zde\v{n}ka Ure\v{s}ov{\'a}. 2015.
\newblock {S}em{E}val 2015 task 18: Broad-coverage semantic dependency parsing.
\newblock In {\em Proc. of SemEval\/}. pages 915--926.

\bibitem[{Oepen et~al.(2014)Oepen, Kuhlmann, Miyao, Zeman, Flickinger,
  Haji\v{c}, Ivanova, and Zhang}]{oepen2014semeval}
Stephan Oepen, Marco Kuhlmann, Yusuke Miyao, Daniel Zeman, Dan Flickinger, Jan
  Haji\v{c}, Angelina Ivanova, and Yi~Zhang. 2014.
\newblock {S}em{E}val 2014 task 8: Broad-coverage semantic dependency parsing.
\newblock In {\em Proc. of SemEval\/}. pages 63--72.

\bibitem[{Pollard and Sag(1994)}]{PandS:94}
Carl Pollard and Ivan Sag. 1994.
\newblock {\em Head Driven Phrase Structure Grammar\/}.
\newblock CSLI Publications, Stanford,~CA.

\bibitem[{Pourdamghani et~al.(2014)Pourdamghani, Gao, Hermjakob, and
  Knight}]{pourdamghani2014aligning}
Nima Pourdamghani, Yang Gao, Ulf Hermjakob, and Kevin Knight. 2014.
\newblock Aligning {E}nglish strings with abstract meaning representation
  graphs.
\newblock In {\em Proc. of EMNLP\/}. pages 425--429.

\bibitem[{Pust et~al.(2015)Pust, Hermjakob, Knight, Marcu, and
  May}]{pust2015parsing}
Michael Pust, Ulf Hermjakob, Kevin Knight, Daniel Marcu, and Jonathan May.
  2015.
\newblock Parsing {E}nglish into abstract meaning representation using
  syntax-based machine translation.
\newblock In {\em Proc. of EMNLP\/}. pages 1143--1154.

\bibitem[{Ribeyre et~al.(2014)Ribeyre, Villemonte de~la Clergerie, and
  Seddah}]{ribeyre-villemontedelaclergerie-seddah:2014:SemEval}
Corentin Ribeyre, Eric Villemonte de~la Clergerie, and Djam\'{e} Seddah. 2014.
\newblock Alpage: Transition-based semantic graph parsing with syntactic
  features.
\newblock In {\em Proc. of SemEval\/}. pages 97--103.

\bibitem[{Roth and Frank(2015)}]{roth2015inducing}
Michael Roth and Anette Frank. 2015.
\newblock {Inducing Implicit Arguments from Comparable Texts: A Framework and
  its Applications}.
\newblock {\em Computational Linguistics\/} 41:625--664.

\bibitem[{Sagae and Lavie(2005)}]{sagae2005classifier}
Kenji Sagae and Alon Lavie. 2005.
\newblock A classifier-based parser with linear run-time complexity.
\newblock In {\em Proc. of IWPT\/}. pages 125--132.

\bibitem[{Sagae and Tsujii(2008)}]{sagae2008shift}
Kenji Sagae and Jun'ichi Tsujii. 2008.
\newblock Shift-reduce dependency {DAG} parsing.
\newblock In {\em Proc. of COLING\/}. pages 753--760.

\bibitem[{Schluter et~al.(2014)Schluter, S{\o}gaard, Elming, Hovy, Plank,
  Mart\'{i}nez~Alonso, Johanssen, and Klerke}]{schluter-EtAl:2014:SemEval}
Natalie Schluter, Anders S{\o}gaard, Jakob Elming, Dirk Hovy, Barbara Plank,
  H\'{e}ctor Mart\'{i}nez~Alonso, Anders Johanssen, and Sigrid Klerke. 2014.
\newblock Copenhagen-{M}alm\"{o}: Tree approximations of semantic parsing
  problems.
\newblock In {\em Proc. of SemEval\/}. pages 213--217.

\bibitem[{Schneider et~al.(2014)Schneider, Danchik, Dyer, and
  Smith}]{schneider2014discriminative}
Nathan Schneider, Emily Danchik, Chris Dyer, and Noah~A Smith. 2014.
\newblock Discriminative lexical semantic segmentation with gaps: running the
  {MWE} gamut.
\newblock {\em TACL\/} 2:193--206.

\bibitem[{Steedman(2000)}]{Steedman:00}
Mark Steedman. 2000.
\newblock {\em The Syntactic Process\/}.
\newblock MIT Press, Cambridge, MA.

\bibitem[{Sulem et~al.(2015)Sulem, Abend, and Rappoport}]{sulem2015conceptual}
Elior Sulem, Omri Abend, and Ari Rappoport. 2015.
\newblock Conceptual annotations preserve structure across translations: A
  {F}rench-{E}nglish case study.
\newblock In {\em Proc. of S2MT\/}. pages 11--22.

\bibitem[{Swayamdipta et~al.(2016)Swayamdipta, Ballesteros, Dyer, and
  Smith}]{swayamdipta-EtAl:2016:CoNLL}
Swabha Swayamdipta, Miguel Ballesteros, Chris Dyer, and Noah~A. Smith. 2016.
\newblock Greedy, joint syntactic-semantic parsing with stack {LSTM}s.
\newblock In {\em Proc. of CoNLL\/}. pages 187--197.

\bibitem[{Thomson et~al.(2014)Thomson, O'Connor, Flanigan, Bamman, Dodge,
  Swayamdipta, Schneider, Dyer, and Smith}]{thomson-EtAl:2014:SemEval}
Sam Thomson, Brendan O'Connor, Jeffrey Flanigan, David Bamman, Jesse Dodge,
  Swabha Swayamdipta, Nathan Schneider, Chris Dyer, and Noah~A. Smith. 2014.
\newblock {CMU}: Arc-factored, discriminative semantic dependency parsing.
\newblock In {\em Proc. of SemEval\/}. pages 176--180.

\bibitem[{Tokg{\"o}z and Eryi\u{g}it(2015)}]{tokgoz2015transition}
Alper Tokg{\"o}z and G{\"u}lsen Eryi\u{g}it. 2015.
\newblock Transition-based dependency {DAG} parsing using dynamic oracles.
\newblock In {\em Proc. of ACL Student Research Workshop\/}. pages 22--27.

\bibitem[{Vanderwende et~al.(2015)Vanderwende, Menezes, and
  Quirk}]{vanderwende2015amr}
Lucy Vanderwende, Arul Menezes, and Chris Quirk. 2015.
\newblock An {AMR} parser for {E}nglish, {F}rench, {G}erman, {S}panish and
  {J}apanese and a new {AMR}-annotated corpus.
\newblock In {\em Proc. of NAACL\/}. pages 26--30.

\bibitem[{Wang et~al.(2016)Wang, Pradhan, Pan, Ji, and
  Xue}]{wang-EtAl:2016:SemEval}
Chuan Wang, Sameer Pradhan, Xiaoman Pan, Heng Ji, and Nianwen Xue. 2016.
\newblock {CAMR} at {S}em{E}val-2016 task 8: An extended transition-based amr
  parser.
\newblock In {\em Proc. of SemEval\/}. pages 1173--1178.

\bibitem[{Wang et~al.(2015{\natexlab{a}})Wang, Xue, and
  Pradhan}]{wang-xue-pradhan:2015:ACL-IJCNLP}
Chuan Wang, Nianwen Xue, and Sameer Pradhan. 2015{\natexlab{a}}.
\newblock Boosting transition-based {AMR} parsing with refined actions and
  auxiliary analyzers.
\newblock In {\em Proc. of ACL\/}. pages 857--862.

\bibitem[{Wang et~al.(2015{\natexlab{b}})Wang, Xue, and
  Pradhan}]{wang2015transition}
Chuan Wang, Nianwen Xue, and Sameer Pradhan. 2015{\natexlab{b}}.
\newblock A transition-based algorithm for {AMR} parsing.
\newblock In {\em Proc. of NAACL\/}. pages 366--375.

\bibitem[{Weischedel et~al.(2013)Weischedel, Palmer, Marcus, Hovy, Pradhan,
  Ramshaw, Xue, Taylor, Kaufman, Franchini et~al.}]{weischedel2013ontonotes}
Ralph Weischedel, Martha Palmer, Mitchell Marcus, Eduard Hovy, Sameer Pradhan,
  Lance Ramshaw, Nianwen Xue, Ann Taylor, Jeff Kaufman, Michelle Franchini,
  et~al. 2013.
\newblock Onto{N}otes release 5.0 {LDC2013T19}.
\newblock {\em Linguistic Data Consortium, Philadelphia, PA\/} .

\bibitem[{Werling et~al.(2015)Werling, Angeli, and Manning}]{werling2015robust}
Keenon Werling, Gabor Angeli, and Christopher~D. Manning. 2015.
\newblock Robust subgraph generation improves abstract meaning representation
  parsing.
\newblock In {\em Proc. of ACL\/}. pages 982--991.

\bibitem[{Zhang and Clark(2009)}]{zhang2009transition}
Yue Zhang and Stephen Clark. 2009.
\newblock Transition-based parsing of the {C}hinese treebank using a global
  discriminative model.
\newblock In {\em Proc. of IWPT\/}. Association for Computational Linguistics,
  pages 162--171.

\bibitem[{Zhang and Clark(2011)}]{zhang2011shift}
Yue Zhang and Stephen Clark. 2011.
\newblock Shift-reduce {CCG} parsing.
\newblock In {\em Proc. of ACL\/}. pages 683--692.

\bibitem[{Zhou et~al.(2016)Zhou, Xu, Uszkoreit, Qu, Li, and Gu}]{zhou2016amr}
Junsheng Zhou, Feiyu Xu, Hans Uszkoreit, Weiguang Qu, Ran Li, and Yanhui Gu.
  2016.
\newblock {AMR} parsing with an incremental joint model.
\newblock In {\em Proc. of EMNLP\/}. pages 680--689.

\bibitem[{Zhu et~al.(2013)Zhu, Zhang, Chen, Zhang, and Zhu}]{zhu2013fast}
Muhua Zhu, Yue Zhang, Wenliang Chen, Min Zhang, and Jingbo Zhu. 2013.
\newblock Fast and accurate shift-reduce constituent parsing.
\newblock In {\em Proc. of ACL\/}. pages 434--443.

\end{thebibliography}
\bibliographystyle{acl_natbib}

\newpage

\appendix
\section{Feature Templates}
\label{appendix:features}

\figref{fig:features} presents the feature templates used by \parser{Sparse}.
All feature templates define binary features.
The other classifiers use the same elements listed in the feature templates,
but all categorical features are replaced by vector embeddings,
and all count-based features are replaced by their numeric value.

For some of the features, we used the notion of \textit{head word},
defined by the $h^*$ function (see Appendix~\ref{appendix:conversion}).
While head words are not explicitly represented in the UCCA scheme, these
features prove useful as means of encoding word-to-word relations.

\begin{figure*}[h]
\centering
\begin{adjustbox}{margin=3pt,frame}
\begin{tabular}{l}
{\footnotesize Features from \cite{zhang2009transition}:} \\
\textbf{unigrams} \\
$s_0tde, s_0we, s_1tde, s_1we, s_2tde, s_2we, s_3tde, s_3we,$ \\
$b_0wtd, b_1wtd, b_2wtd, b_3wtd,$ \\
$s_0lwe, s_0rwe, s_0uwe, s_1lwe, s_1rwe, s_1uwe$ \\
\textbf{bigrams} \\
$s_0ws_1w, s_0ws_1e, s_0es_1w, s_0es_1e, s_0wb_0w, s_0wb_0td,$ \\
$s_0eb_0w, s_0eb_0td, s_1wb_0w, s_1wb_0td, s_1eb_0w, s_1eb_0td,$ \\
$b_0wb_1w, b_0wb_1td, b_0tdb_1w, b_0tdb_1td$ \\
\textbf{trigrams} \\
$s_0es_1es_2w, s_0es_1es_2e, s_0es_1eb_0w, s_0es_1eb_0td,$ \\
$s_0es_1wb_0w, s_0es_1wb_0td, s_0ws_1es_2e, s_0ws_1eb_0td$ \\
\textbf{separator} \\
$s_0wp, s_0wep, s_0wq, s_0wcq, s_0es_1ep, s_0es_1eq,$ \\
$s_1wp, s_1wep, s_1wq, s_1weq$ \\

\textbf{extended} \footnotesize \cite{zhu2013fast} \\
$s_0llwe, s_0lrwe, s_0luwe, s_0rlwe, s_0rrwe,$ \\
$s_0ruwe, s_0ulwe, s_0urwe, s_0uuwe, s_1llwe,$ \\
$s_1lrwe, s_1luwe, s_1rlwe, s_1rrwe, s_1ruwe$ \\
\end{tabular}
\begin{tabular}{l}
\textbf{disco} \footnotesize \cite{maier2015discontinuous} \\
$s_0xwe, s_1xwe, s_2xwe, s_3xwe,$ \\
$s_0xtde, s_1xtde, s_2xtde, s_3xtde,$ \\
$s_0xy, s_1xy, s_2xy, s_3xy$ \\
$s_0xs_1e, s_0xs_1w, s_0xs_1x, s_0ws_1x, s_0es_1x,$ \\
$s_0xs_2e, s_0xs_2w, s_0xs_2x, s_0ws_2x, s_0es_2x,$ \\
$s_0ys_1y, s_0ys_2y, s_0xb_0td, s_0xb_0w$ \\

{\footnotesize Features from \cite{tokgoz2015transition}:} \\
\textbf{counts} \\
$s_0P, s_0C, s_0wP, s_0wC, b_0P, b_0C, b_0wP, b_0wC$ \\
\textbf{edges} \\
$s_0s_1, s_1s_0, s_0b_0, b_0s_0, s_0b_0e, b_0s_0e$ \\
\textbf{history} \\
$a_0, a_1$ \\

\textbf{remote} \footnotesize (Novel, UCCA-specific features) \\
$s_0R, s_0wR, b_0R, b_0wR$
\end{tabular}
\end{adjustbox}
\captionsetup{singlelinecheck=off}
\caption[]{\label{fig:features}
  Binary feature templates for \parser{Sparse}. Notation:\\
  $s_i$, $b_i$: $i$th stack and buffer items.\\
  $w$, $t$, $d$: word form, POS tag and syntactic dependency label of the terminal returned by $h^*(\cdot)$
  (see Appendix~\ref{appendix:conversion}).\\
  $e$: edge label to the node returned by $h(\cdot)$.\\
  $l$, $r$ ($ll$, $rr$): leftmost and rightmost (grand)children.\\
  $u$ ($uu$): unary (grand)child, when only one exists.\\
  $p$: unique separator punctuation between $s_0$ and $s_1$. $q$: separator count.\\
  $x$: gap type (``none'', ``pass'' or ``gap'') at the sub-graph under the current node.\\
  $y$: sum of gap lengths \protect\cite{Maier:Lichte:11}.\\
  $P$, $C$: number of parents and children.\\
  $R$: number of remote children.\\
  $a_i$: action taken $i$ steps back.
}
\end{figure*}

\section{Extended Presentation of UCCA}
\label{appendix:extended_ucca}

This work does not handle two important constructions in the UCCA foundational layer:
Linkage, representing discourse relations, and Implicit, representing covert entities.
\tabref{table:data_linkage_implicit} shows the statistics of linkage nodes and edges and
implicit nodes in the corpora.

\begin{table}[h]
\centering
\begin{tabular}{l|ccc|c}
& \multicolumn{3}{c|}{Wiki} & 20K \\
& \small Train & \small Dev & \small Test & Leagues \\
\hline
nodes \\
\# implicit & 899 & 122 & 77 & 241 \\
\# linkage & 2956 & 263 & 359 & 376 \\
\hline
edges \\
\# linkage & 9276 & 803 & 1094 & 957
\end{tabular}
\caption{Statistics of linkage and implicit nodes in the
\textit{Wiki} and \textit{20K Leagues} UCCA corpora.
Cf. \tabref{table:data}.
}
\label{table:data_linkage_implicit}
\end{table}

\paragraph{Linkage.}

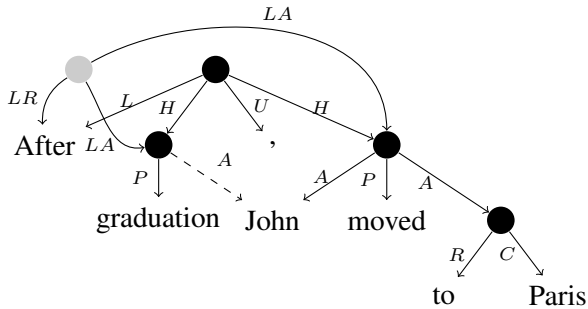
\begin{figure}[h]
  \centering
  \begin{tikzpicture}[level distance=10mm, ->]
    \node (ROOT) [fill=black, circle] {}
      child {node (After) {After} edge from parent node[left] {\scriptsize $L$}}
      child {node (graduation) [fill=black, circle] {}
      {
        child {node {graduation} edge from parent node[left] {\scriptsize $P$}}
      } edge from parent node[left] {\scriptsize $H$} }
      child {node {,} edge from parent node[right] {\scriptsize $U$}}
      child {node (moved) [fill=black, circle] {}
      {
        child {node (John) {John} edge from parent node[left] {\scriptsize $A$}}
        child {node {moved} edge from parent node[left] {\scriptsize $P$}}
        child {node [fill=black, circle] {}
        {
          child {node {to} edge from parent node[left] {\scriptsize $R$}}
          child {node {Paris} edge from parent node[left] {\scriptsize $C$}}
        } edge from parent node[left] {\scriptsize $A$} }
      } edge from parent node[right] {\scriptsize $H$} }
      ;
    \draw[dashed,->] (graduation) to node [auto] {\scriptsize $A$} (John);
    \node (LKG) at (-1.8,0) [fill=black!20, circle] {};
          \draw[bend right] (LKG) to node [auto, left] {\scriptsize $LR$} (After);
          \draw (LKG) to[out=-60, in=190] node [below] {\scriptsize $LA\quad$} (graduation);
          \draw (LKG) to[out=30, in=90] node [above] {\scriptsize $LA$} (moved);
  \end{tikzpicture}
  \caption{UCCA example with linkage.}
  \label{fig:example_linkage}
\end{figure}

\figref{fig:example_linkage} demonstrates a linkage relation, omitted from \figref{fig:graduation}.
The linkage relation is represented by the gray node.
$LA$ is \emph{link argument}, and $LR$ is \emph{link relation}.
The relation represents the fact that the \emph{linker} ``After'' links the two parallel scenes
that are the arguments of the linkage.
Linkage relations are another source of multiple parents for a node, which we do not yet handle
in parsing and evaluation.

\paragraph{Implicit units.}

\begin{figure*}[h]
  \centering
  \scalebox{.6}{
  \begin{tikzpicture}[level distance=20mm, ->,
  level 1/.style={sibling distance=8em},
  level 2/.style={sibling distance=4em},
  level 3/.style={sibling distance=4em}]
    \node (ROOT) [fill=black, circle] {}
      child {node [fill=black, circle] {}
      {
        child {node {A} edge from parent node[left] {\scriptsize $E$}}
        child {node {similar} edge from parent node[left] {\scriptsize $E$}}
        child {node {technique} edge from parent node[right] {\scriptsize $C$}}
      } edge from parent node[left] {\scriptsize $A\quad$ \hspace{1mm} } }
      child {node {is} edge from parent node[left] {\scriptsize $F$}}
      child {node [fill=black, circle] {}
      {
        child {node {almost} edge from parent node[left] {\scriptsize $E$}}
        child {node {impossible} edge from parent node[right] {\scriptsize $C$}}
      } edge from parent node[left] {\scriptsize $D$} }
      child {node {\textbf{IMPLICIT}} edge from parent node[left] {\scriptsize $A$}}
      child {node {to} edge from parent node[left] {\scriptsize $F$}}
      child {node {apply} edge from parent node[left] {\scriptsize $P\quad$}}
      child {node [fill=black, circle] {}
      {
        child {node {to} edge from parent node[left] {\scriptsize $R$}}
        child {node {other} edge from parent node[left] {\scriptsize $E$}}
        child {node {crops} edge from parent node[left] {\scriptsize $C$}}
        child {node {,} edge from parent node[left] {\scriptsize $U$}}
        child {node [fill=black, circle] {}
        {
          child {node {such as} edge from parent node[left] {\scriptsize $R$}}
          child {node {cotton} edge from parent node[left] {\scriptsize $C$}}
          child {node {,} edge from parent node[left] {\scriptsize $U$}}
          child {node {soybeans} edge from parent node[left] {\scriptsize $C$}}
          child {node {and} edge from parent node[left] {\scriptsize $N$}}
          child {node {rice} edge from parent node[right] {\scriptsize $\; C$}}
        } edge from parent node[right] {\scriptsize $\; E$ \hspace{1mm} } }
      } edge from parent node[left] {\scriptsize $A\;$ \hspace{1mm} } }
      child {node {.} edge from parent node[right] {\scriptsize $\quad \quad U$}}
      ;
  \end{tikzpicture}
  }
  \caption{UCCA example with an implicit unit.}
  \label{fig:example_implicit}
\end{figure*}
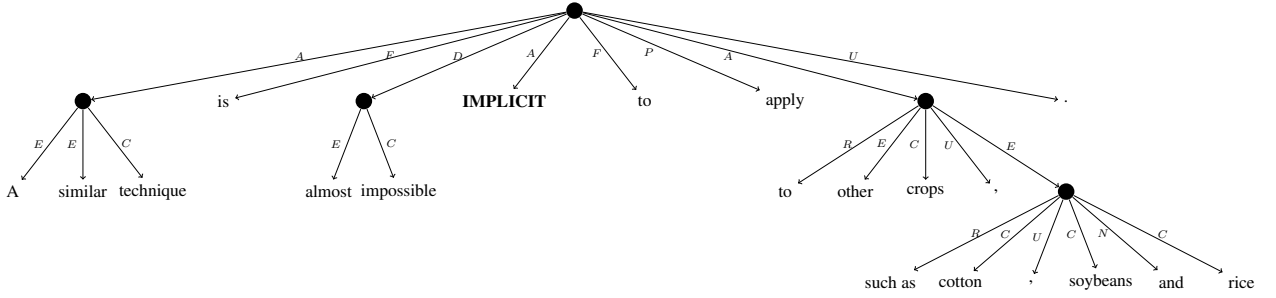

UCCA graphs may contain implicit units with no correspondent in the text.
\figref{fig:example_implicit} shows the annotation for the sentence
``A similar technique is almost impossible to apply to other crops, such as cotton, soybeans and rice.''.
The sentence was used by \newcite{oepen2015semeval} to compare between different semantic
dependency schemes.
It includes a single scene, whose main relation is ``apply'', a secondary relation ``almost impossible'', as well as two complex arguments: ``a similar technique'' and the coordinated argument ``such as cotton, soybeans, and rice.''
In addition, the scene includes an implicit argument, which represents the agent of the
``apply'' relation.

The parsing of these units is deferred to future work, as it is likely to require different methods
than those explored in this paper \cite{roth2015inducing}.

\section{Hyperparameter Values}
\label{appendix:hyperparameters}

\tabref{table:hyperparameters} lists the hyperparameter values we found
for the different classifiers by tuning on the development set.
Note that learning rate decay is multiplicative and is applied at each epoch.
Mini-batch size is in number of transitions,
but a mini-batch must contain only whole sentences.

\begin{table}[h]
\centering
\scalebox{.9}{
\begin{tabular}{l|ccc}
& Sparse & MLP & BiLSTM \\
\hline
\multicolumn{4}{l}{\footnotesize Embedding dimensions} \\
external word & & 100 & 100 \\
word & & 200 & 200 \\
POS tag & & 20 & 20 \\
syntactic dep. & & 10 & 10 \\
edge label & & 20 & 20 \\
punctuation & & 1 & 1 \\
gap & & 3 & 3 \\
action & & 3 & 3 \\
\hline
\multicolumn{4}{l}{\footnotesize Other parameters} \\
training epochs & 19 & 28 & 59 \\
$\textsc{MinUpdate}$ & 5 \\
initial learning rate & 1 & 1 & 1 \\
learning rate decay & 0.1 & 1 & 1 \\
MLP \#layers & & 2 & 2 \\
MLP layer dim. & & 100 & 50 \\
LSTM \#layers & & & 2 \\
LSTM layer dim. & & & 500 \\
word dropout & & 0.2 & 0.2 \\
dropout & & 0.4 & 0.4 \\
weight decay & & $10^{-5}$ & $10^{-5}$ \\
mini-batch size & & 100 & 100
\end{tabular}
}
\caption{Hyperparameters used for the different classifiers.}
\label{table:hyperparameters}
\end{table}

\section{Bilexical Graph Conversion}
\label{appendix:conversion}

Here we describe the algorithms used in the conversion referred to in \secref{sec:exp_setup}.

\paragraph{Notation.}
Let $L$ be the set of possible edge labels.
A UCCA graph over a sequence of tokens $w_1, \ldots, w_n$ is a directed acyclic graph
$G=(V,E, \ell)$, where $\ell:E\to L$ maps edges to labels.
For each token $w_i$ there exists a leaf (\emph{terminal}) $t_i \in V$.
A bilexical (dependency) graph over the same text consists of a set $A$ of
labeled dependency arcs $(t^\prime,l,t)$
between the terminals of $G$, where $t^\prime$ is the head, $t$ is the dependent and $l$ is
the edge label.

\paragraph{Conversion to bilexical graphs.}
Let $G=(V,E,\ell)$ be a UCCA graph with labels $\ell:E\rightarrow L$.
The conversion to a bilexical graph requires calculating the set $A$.
All non-terminals in $G$ are removed.

We define a linear order over possible edge labels $L$ (see \figref{fig:priority}).
The priority order generally places core-like categories before adjunct-like ones, and was decided heuristically.
For each node $u \in V$, denote by $h(u)$ its child with the highest-priority edge label.
The leftmost edge is chosen in case of a tie.
Let $h^*(u)$ be the terminal reached by recursively applying $h(\cdot)$ over $u$.
For each terminal $t$, we define
\[
N(t) = \{(u,v)\in E \;|\; t=h^*(v) \wedge t \neq h^*(u) \}
\]
For each edge $(u,v)\in N(t)$, we add $h^*(u)$ as a head of $t$ in $A$,
with the label $\ell(u,v)$.
This procedure is given in Algorithm~\ref{alg:to_bilexical}.

\begin{algorithm}[ht]
 \KwData{UCCA graph ${G}=(V,E,\ell)$}
 \KwResult{set $A$ of labeled bilexical arcs}
 $A \leftarrow \emptyset$\;
 \ForEach{$t \in \mathrm{Terminals}(V)$} {
  \ForEach{$(u,v)\in N(t)$} {
   $A \leftarrow A \cup \{(h^*(u), \ell(u, v), t)\}$\;
  }
 }
 \caption{Conversion to bilexical graphs.}
 \label{alg:to_bilexical}
\end{algorithm}

Note that this conversion procedure
is simpler than the head percolation procedure used for converting syntactic constituency
trees to dependency trees \cite{Coll:97},
since $h(u)$ (similar to $u$'s head-containing child)
depends only on $\ell(u,h(u))$ and not on the sub-tree spanned by $u$,
because edge labels in UCCA directly express the role of the child in the parent unit, and
are thus sufficient for determining which of $u$'s children contains the head node.

\paragraph{Conversion from bilexical graphs.}
The inverse conversion introduces non-terminal nodes back into the graph.
As the distinction between low- and high-attaching nodes is lost in the
conversion, we assume that attachments are always
low-attaching.
Let $A$ be a the labeled arc set of a bilexical graph.
Iterating over the terminals in topological order according to $A$,
we add its members as terminals to graph
and create a pre-terminal parent $u_t$ for each terminal $t$,
with an edge labeled as \textit{Terminal} between them.
The parents of the pre-terminals are determined by the terminal's parent in the bilexical
graph: if $t^\prime$ is a head of $t$ in $A$, then $u_{t^\prime}$ will be a parent of $u_t$.
We add an intermediate node in between if $t$ has any dependents in $A$,
to allow adding their pre-terminals as children later.
Edge labels for the intermediate edges are determined by a rule-based function, denoted by
$\mathrm{Label}(t)$.
This procedure is given in Algorithm~\ref{alg:from_bilexical}.

\begin{algorithm}[ht]
 \KwData{list $T$ of terminals, set $A$ of labeled bilexical arcs}
 \KwResult{UCCA graph $G=(V,E,\ell)$}
 \SetKwFunction{Label}{Label}{}{}
 $V \leftarrow \emptyset$,
 $E \leftarrow \emptyset$\;
 \ForEach{$t \in \mathrm{TopologicalSort}(T, A)$} {
  $u_t \leftarrow \mathrm{Node()}$\;
  $V \leftarrow V \cup \{u_t, t\}$,
  $E \leftarrow E \cup \{(u_t, t)\}$\;
  $\ell(u_t,t)\leftarrow\mathit{Terminal}$\;
  \ForEach{$t^\prime\in T,l\in L$} {
   \If{$(t^\prime,l,t)\in A$} {
    \eIf{$\exists t^{\prime\prime}\in T,l^\prime\in L : (t,l^\prime,t^{\prime\prime}) \in A$} {
     $u \leftarrow \mathrm{Node()}$\;
     $V \leftarrow V \cup \{u\}$,
     $E \leftarrow E \cup \{(u, u_t)\}$\;
     $\ell(u, u_t) \leftarrow \Label(t)$\;
    } {
     $u \leftarrow u_t$\;
    }
    $E \leftarrow E \cup \{(u_{t^\prime}, u)\}$\;
    $\ell(u_{t^\prime}, u) \leftarrow l$\;
    }
  }
 }
 
  \SetKwProg{func}{Function}{}{}
  
  \func{\Label}{
  \KwData{node $t \in T$}
  \KwResult{label $l\in L$}

   \uIf{$\mathrm{IsPunctuation}(t)$}{
    \Return \textit{Punctuation}\;
   }
   \uElseIf{$\exists t^\prime \in T : (t,\textit{ParallelScene},t^\prime)\in A$}{
    \Return \textit{ParallelScene}\;
   }
   \uElseIf{$\exists t^\prime \in T : (t,\textit{Participant},t^\prime)\in A$}{
    \Return \textit{Process}\;
   }
   \uElse{
    \Return \textit{Center}\;
   }
  }
 \caption{Conversion from bilexical graphs.}
 \label{alg:from_bilexical}
\end{algorithm}

\begin{figure}[ht]
\begin{multicols}{2}
\begin{enumerate}
\itemsep0em
\item $C$ (Center)
\item $N$ (Connector)
\item $H$ (ParallelScene)
\item $P$ (Process)
\item $S$ (State)
\item $A$ (Participant)
\item $D$ (Adverbial)
\item $T$ (Time)
\item $E$ (Elaborator)
\item $R$ (Relator)
\item $F$ (Function)
\item $L$ (Linker)
\item $LR$ (LinkRelation)
\item $LA$ (LinkArgument)
\item $G$ (Ground)
\item $\mathit{Terminal}$ (Terminal)
\item $U$ (Punctuation)
\end{enumerate}
\end{multicols}
\caption{Priority order of edge labels used by $h(u)$.}
\label{fig:priority}
\end{figure}

\section{Proof Sketch for Completeness of the \parser{} Transition Set}
\label{appendix:completeness_proof}

Here we sketch a proof for the fact that the transition set defined in \secref{sec:parser}
is capable of producing any rooted, labeled, anchored DAG.
This proves that the transition set is complete with respect to the class of graphs that
comprise UCCA.

Let $G=(V,E,\ell)$ be a graph with labels $\ell:E\rightarrow L$
over a sequence of tokens $w_1, \ldots, w_n$.
Parsing starts with $w_1, \ldots, w_n$ on the buffer,
and the root node on the stack.

First we show that every node can be created, by induction on the node height:
every terminal (height zero) already exists at the beginning of the parse
(and so does the root node).
Let $v\in V$ be of height $k$, and assume all nodes of height less than $k$ can be created.
Take any (primary) child $u$ of $v$: its height must be less than $k$.
If $u$ is a terminal, apply \textsc{Shift} until it lies at the head of the buffer.
Otherwise, by our assumption, $u$ can still be created.
Right after $u$ is created, it lies at the head of the buffer.
A \textsc{Shift} transition followed by a \textsc{Node}$_{\ell(v,u)}$ transition will
move $u$ to the stack and create $v$ on the buffer, with the correct edge label.

Next, we show that every edge can be created.
Let $(v,u) \in E$ be any edge with parent $v$ and child $u$.
Assume $v$ and $u$ have both been created (we already showed that both are created eventually).
If either $v$ or $u$ are in the buffer, apply \textsc{Shift} until both are in the stack.
If both are in the stack but neither is at the stack top, apply \textsc{Swap} transitions
until either moves to the buffer, and then apply \textsc{Shift}.
Now, assume either $v$ or $u$ is at the stack top.
If the other is not the second element on the stack, apply \textsc{Swap} transitions until it is.
Finally, $v$ and $u$ are the top two elements on the stack.
If they are in that order, apply \textsc{Right-Edge}$_{\ell(v,u)}$
(or \textsc{Right-Remote}$_{\ell(v,u)}$ if the edge between them is remote).
Otherwise, apply \textsc{Left-Edge}$_{\ell(v,u)}$
(or \textsc{Left-Remote}$_{\ell(v,u)}$ if the edge between them is remote).
This creates $(v,u)$ with the correct edge label.

Once all nodes and edges have been created, we can apply \textsc{Reduce} until only the
root node remains on the stack, and then \textsc{Finish}.
This yields exactly the graph $G$.

Note that the distinction we made between primary and remote transitions is suitable for UCCA parsing.
For general graph parsing without this distinction,
the \textsc{Remote} transitions can be removed, as well as the single-primary-parent
restriction on \textsc{Edge} transition.

\end{document}